\documentclass[acmsmall,screen]{acmart}
\usepackage{multirow}
\usepackage{multicol}
\usepackage{longtable}
\usepackage{dsfont}
\usepackage{mathtools}
\usepackage{tikz}
\usepackage{colortbl}
\usepackage{microtype}
\usepackage{ltablex}
\usepackage{tabularx}
\usepackage{longtable}

\usepackage{xcolor}

\definecolor{mygreen}{RGB}{14, 107, 14}
\definecolor{myred}{RGB}{151, 0, 0}
\definecolor{myBlue}{RGB}{214, 234, 248}
\definecolor{myWhite}{RGB}{255, 255, 255}
\definecolor{myOrange}{RGB}{250, 219, 216}

\newcommand{\greencheck}{}%
\DeclareRobustCommand{\greencheck}{%
  \tikz\fill[scale=0.45, color=mygreen]
  (0,.35) -- (.25,0) -- (1,.7) -- (.25,.15) -- cycle;%
}

\newcolumntype{?}{!{\vrule width 1.75pt}}

\newcommand{\crosscheck}{$\mathbin{\tikz[myred, line width=.4ex ] \filldraw[color=myred] (0,0) circle (.5ex);}$}%

\AtBeginDocument{%
  }

\begin{document}


\title{Benchmarking Instance-Centric Counterfactual Algorithms for XAI: From White Box to Black Box}

\author{Catarina Moreira}
\email{catarina.pintomoreira@uts.edu.au}
\orcid{0000-0002-8826-5163}
\additionalaffiliation{
  \institution{INESC-ID Lisboa / Instituto Superior T\'{e}cnico, ULisboa}
  \streetaddress{Avenida Professor Cavaco Silva, Edif\'{i}cio IST}  
  \city{Oeiras}
  \country{Portugal}
  \postcode{2744-016 Porto Salvo}
 }
 
 \affiliation{
  \institution{Human Technology Institute, University of Technology Sydney}
  \streetaddress{15 Broadway, Ultimo}
  \city{Sydney}
  \state{New South Wales}
  \country{Australia}
  \postcode{2007}
}

\author{Yu-Liang Chou}
\email{yuliang.chou@hdr.qut.edu.au}
\orcid{0000-0001-9881-953X}
\affiliation{
  \institution{School of Information Systems, Queensland University of Technology}
  \streetaddress{2 George Street}
  \city{Brisbane}
  \state{Queensland}
  \country{Australia}
  \postcode{4000}
}

\author{Chihcheng Hsieh}
\email{chihcheng.hsieh@hdr.qut.edu.au}
\orcid{0000-0002-3352-6899}

\affiliation{
  \institution{School of Information Systems, Queensland University of Technology}
  \streetaddress{2 George Street}
  \city{Brisbane}
  \state{Queensland}
  \country{Australia}
  \postcode{4000}
}

\author{Chun Ouyang}
\orcid{0000-0001-7098-5480}
\affiliation{%
  \institution{School of Information Systems, Queensland University of Technology}
  \streetaddress{2 George Street}
  \city{Brisbane}
  \state{Queensland}
  \country{Australia}
  \postcode{4000}
}
\email{c.ouyang@qut.edu.au}

\author{Jo\~{a}o Madeiras Pereira}
\orcid{0000-0002-8120-7649}
\affiliation{%
  \institution{INESC-ID Lisboa / Instituto Superior T\'{e}cnico, ULisboa}
  \streetaddress{Avenida Professor Cavaco Silva, Edif\'{i}cio IST}  
  \city{Oeiras}
  \country{Portugal}
  \postcode{2744-016 Porto Salvo}
 }
 \email{jap@inesc-id.pt}

\author{Joaquim Jorge}
\orcid{0000-0001-5441-4637}
\affiliation{%
  \institution{INESC-ID Lisboa / Instituto Superior T\'{e}cnico, ULisboa}
  \streetaddress{Avenida Professor Cavaco Silva, Edif\'{i}cio IST}  
  \city{Oeiras}
  \country{Portugal}
  \postcode{2744-016 Porto Salvo}
  }
  \email{jorgej@acm.org}

\renewcommand{\shortauthors}{Moreira et al.}

\begin{abstract}
\textbf{ABSTRACT}\\
This study investigates the impact of machine learning models on the generation of counterfactual explanations by conducting a benchmark evaluation over three different types of models: a decision tree (fully transparent, interpretable, white-box model), a random forest (semi-interpretable, grey-box model), and a neural network (fully opaque, black-box model). We tested the counterfactual generation process using four algorithms (DiCE, WatcherCF, prototype, and GrowingSpheresCF) in the literature in 25 different datasets.
Our findings indicate that:
(1) Different machine learning models have little impact on the generation of counterfactual explanations;
(2) Counterfactual algorithms based uniquely on proximity loss functions are not actionable and will not provide meaningful explanations;
(3) One cannot have meaningful evaluation results without guaranteeing plausibility in the counterfactual generation. Algorithms that do not consider plausibility in their internal mechanisms will lead to biased and unreliable conclusions if evaluated with the current state-of-the-art metrics;
(4) A counterfactual inspection analysis is strongly recommended to ensure a robust examination of counterfactual explanations and the potential identification of biases. \\


\end{abstract}


\maketitle

\section{Introduction} \label{sec:introduction}

The rapidly growing adoption of Artificial intelligence (AI) has led to the development of deep neural networks for high predictive accuracy~\cite{Chen2021,Murdoch19} in recent years. This advancement has significantly improved the state of the art in many fields, including computer vision, speech recognition, e-commerce, banking, healthcare, etc.~\cite{Mothilal2021,hsieh2023mdf,neves2024shedding}. Although advanced machine learning techniques are widely applied in industry, their sophisticated underlying mechanisms are opaque and do not give the user any understanding of their internal predictive mechanisms.
This opaqueness results in several issues, including fairness, accountability, and transparency, which may violate government regulations (e.g., the General Data Protection Regulation (GDPR))~\cite{Goodman2017,gunning2019}. The ambiguity in machine learning models (ML) is known as \textit{the black box problem}. It is hard for a user to understand why a particular prediction was made, consequently generating a lack of trust in the model. 

The black box problem has drawn the attention of researchers who are trying to understand \textit{why} and \textit{how} an AI system produces a specific outcome or forecast in a research field called Explainable Artificial Intelligence~\cite{Keane2021, Moreira2020icsoc,Moreira2020bpm}. 
\textit{Explainability} is a term that refers to the set of methods that allows human users to comprehend and trust the results and output created by machine learning algorithms. Explainable AI describes the expected impact and potential biases of an AI model. It can also help models comply with legal requirements and increase model reliability. A thorough and precise account of how a model generated its outcome is what we refer to as an explanation~\cite{Alikhademi2021}. It is important to note that trust in AI involves two major components: explainability and robustness. Explainability provides insight into why a model has arrived at a specific output, contributing to the model's transparency. On the other hand, robustness ensures that even with small perturbations in the input data, the model remains consistent in its predictions. This robustness is particularly crucial in domains with poor data quality, where idealized data assumptions may not hold. A comprehensive treatment of both these aspects fosters a truly trustworthy AI. However, neural networks, one of the most powerful learning algorithms, often lack both traits, which motivates the study of tools such as counterfactual explanations~\cite{Holzinger2022Fusion, Holzinger2021next}.

\subsection{Counterfactual Explanations}
\begin{figure}[!ht]
    \centering
    \resizebox{\columnwidth}{!} {
    \includegraphics[scale=0.75]{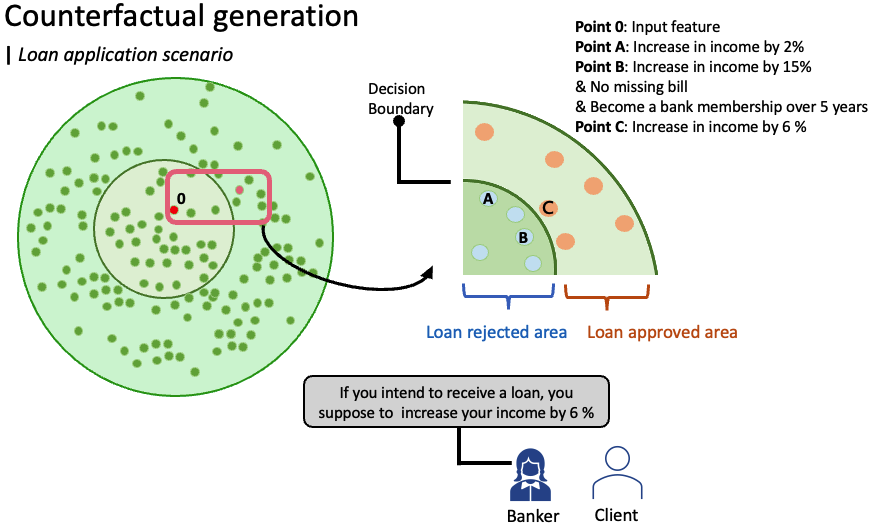}
    }
    \caption{Counterfactual generation graph: Each point in the graph contains a different condition of an applicant, including age, credit amount, credit history, etc.
    A generated counterfactual contains the different conditions from the input of an applicant (point 0) that lead to a loan-approved result (orange dots). For instance, point $c$ indicates the applicant could be granted a loan if the income could be increased by 6\%. Points $A$ and $B$ are the conditions for the applicant to remain loan rejected.}
    \label{fig:cf_generation_pineline}
\end{figure}

Recently, counterfactual explanations are considered an important post-hoc method that gives persuasive explanations for users to understand the internal mechanisms of AI models~\cite{Miller2019,Biran2017survey, Byrne2019,Watcher2018,Guidotti2019}. Unlike scoring or feature attribution explanation methods, which express each feature's (relative) relevance to the model's output~\cite{Mothilal2021}, counterfactual explanations show which modifications would be required to get the desired result. This implies that the counterfactual generation process \textit{is resumed to an optimization problem} where the change between the original query and the candidate counterfactual with the desired outcome is the minimum possible. This technique is described as a conditional assertion with a false antecedent and a consequent that depicts how the world would have been if the antecedent had happened (a what-if question)~\cite{Lewis1973}. For example, in a scenario where a machine learning algorithm determines whether a person should be granted a loan or not, a counterfactual explanation of why a person was denied a loan may be in the form of a scenario in which \textit{you would have been awarded a loan if your income had been more than $8,000$ a year} (Figure~\ref{fig:cf_generation_pineline}).   

Trust in AI systems is crucial and can be significantly enhanced through counterfactual explanations. These explanations play a key role by elucidating the conditions under which different outcomes would occur, helping users understand and navigate AI's otherwise opaque decision-making processes~\cite{del2024generating}. By exploring hypothetical modifications that would lead to an alternate result, counterfactual explanations foster familiarity with AI systems and enhance trust by making these systems more transparent and comprehensible than feature attribution methods \cite{alzubaidi2023towards,longo2024explainable}.

\subsection{The Problem of Validation of Counterfactual Explanations}

Although counterfactuals have been recently explored in the literature, they lack principled approaches and standardized protocols for evaluation. This could be because researchers focus on creating counterfactuals by utilizing different optimization approaches and heuristic rules to find the minimum change that would lead to the desired outcome. These approaches change significantly with different counterfactual models. So, there is no consistent way of finding this minimum counterfactual. Although some metrics are used for feature attribution XAI algorithms, such as fidelity~\cite{Moreira2021fidelity}, and stability~\cite{Moreira2021stability}, there is no standardized way of evaluating XAI algorithms in general, which increases the complexity and difficulty of developing a benchmark evaluation for counterfactuals~\cite{Bordt2022,velmurugan2023through}.

Another important open question in the literature is how different types of machine learning algorithms affect the generation of counterfactuals. Are counterfactuals generated by a deep neural network more difficult to find or interpret, given the complexity of its internal mechanisms? Or are they easier to interpret given a grey model such as a Random Forest classifier? Or are they even easier to find in a white box model such as a decision tree?

To our knowledge, the publication most closely aligned with these research gaps was recently proposed by~\cite{Mazzine2021survey,Guidotti2022}. In their paper, the authors provide a benchmark for representative counterfactuals in the literature for different evaluation metrics. However, the impact of different machine learning models on the counterfactual generation process was not investigated. A thorough analysis of the different counterfactual loss functions and their impact on the different quantitative metrics was also missing.

This work extends the counterfactual evaluation process by including a counterfactual inspection analysis. We argue that relying blindly on quantitative evaluation metrics without understanding the different properties of the counterfactual algorithms may lead to biased and erroneous explanations. We argue that an evaluation protocol for XAI counterfactuals should cover not only quantitative metrics but also a counterfactual inspection that assesses the generated counterfactuals towards the context of the data and the domain knowledge of the decision-maker. For instance, in Hsieh et al.~\cite{Moreira2021icpm}, the authors applied a counterfactual algorithm, DiCE, to generate explanations for a supervised model that predicted the next activity of a loan application process. Although DiCE~\cite{Mothilal2020DiCE} could generate counterfactuals with the minimum feature change, it could not generate meaningful or interpretable counterfactuals to the loan application process (domain knowledge). For this reason, the authors proposed an extension of DiCE that could consider the process domain knowledge and consequently generate more feasible and meaningful counterfactual explanations. This suggests that relying uniquely on quantitative measures does not guarantee the correctness of the generated counterfactual explanations, and blindly relying on quantitative metrics to assess the quality of a counterfactual explanation may lead to biased and unreliable scientific conclusions~\cite{Kim2016}. Unfortunately, this is the direction that most counterfactual studies for XAI take in the literature~\cite{Das2020survey,Biran2017survey,Guidotti2019survey,Mazzine2021survey,Karimi2021survey}.

Given that the formalization of a counterfactual benchmark evaluation is still in its early stages, to the best of our knowledge, none of the existing studies makes a deep analysis of the impact of various machine learning algorithms on the counterfactual generation process, and none of them investigates biases in the counterfactual generation process. Hence, we propose three research questions: (1) Are the present counterfactual evaluation metrics sufficient to measure the quality of the generated counterfactual explanations? (2) Does the choice of machine learning algorithm affect the counterfactual generation process? (3) What contextual mechanisms can we use to find biases in the counterfactual generation process (or predictive model) to assess the counterfactuals' alignment with context and domain knowledge? Answering these three research questions is the aim of this work. To answer it, we (1) Provide a comprehensive evaluation benchmark over several quantitative state-of-the-art metrics; (2) Compare the performance of the counterfactual generation process with the different machine learning models (a white box, a grey box, and a black box), and we (3) Perform a counterfactual inspection analysis where we investigate how counterfactuals are generated in a decision tree. Note that by \textit{bias}, we refer to the potential systematic favorability or unfavorability that these XAI algorithms might exhibit towards certain counterfactuals. In other words, these algorithms might consistently generate counterfactuals that are, for example, closer to the query data point at the expense of overlooking the plausibility of these counterfactuals.

\subsection{Contribution}

The main contributions of this work are the following:
\begin{enumerate}
    \item We explore the capability of instance-centric counterfactual explanations ~\cite{Chou2022fusion}: DiCE, Prototype, GrowingSpheresCf and WatcherCF.
    
    \item We investigate the impact of adopting different machine learning models on four selective instances-centric counterfactual algorithms.
    
    \item We propose a benchmark evaluation of the properties of each counterfactual algorithm, such as proximity, interpretability, and functionality. This benchmark framework implementation assesses different counterfactual generation algorithms. The framework is extendable, allowing for the easy addition of new algorithms, and it may be used to evaluate and compare other counterfactual-generating algorithms. It is open source, and the experiments can be found in \url{https://github.com/LeonChou5311/Counterfactual-benchmark}.
    
    \item We propose adding a counterfactual inspection of counterfactual explanations by analyzing the decision paths between the input vector and its respective counterfactuals. This analysis can provide insights into the counterfactual explanation process, identify potential biases, and provide insights on how to generate better counterfactuals.
\end{enumerate}

\subsection{Findings}
Our experiments revealed that:
\begin{enumerate}
    \item Relying solely on quantitative metrics, such as proximity or sparsity, is insufficient and a poor indicator of assessing the quality of a counterfactual explanation;
    
    \item Explainable counterfactual algorithms that do not take into consideration plausibility in their internal mechanisms cannot be evaluated with 
    state-of-the-art evaluation metrics and their results may be biased and lead to scientific misinterpretations;
    
    \item A counterfactual inspection analysis is strongly recommended (together with a quantitative analysis) to ensure a robust analysis of counterfactual explanations. Generating counterfactuals' decision paths can provide novel insights into the counterfactual generation process, provide a plausibility analysis, or even verify alignment with context and domain knowledge. This cannot be achieved with the current quantitative metrics.
    
    \item The underlying predictive model (either a white box, grey box, or black box) has no significant impact on the counterfactual generation process.
    
    \item DiCE achieved the best overall results because it satisfies the plausibility property, and GrowingSpheresCF achieved the best outcomes regarding proximity and sparsity. On the other hand, WatcherCF achieved the worst results, and we do not recommend the usage of this algorithm in an explainable system since it can easily lead to biased outcomes.
    
\end{enumerate}

\section{Background and Related Work}

\subsection{Background}

Various approaches have been proposed in the literature to address the problem of interpretability~\cite{Burkart2021,Guidotti2019survey}. In general, explainable models can be categorized into two main approaches: transparent and opaque models~\cite{Belle2021, Fernandez2019}. Transparent models are already interpretable by design. They allow 
people to understand how the model works by directly inspecting and extracting its feature importance. Decision trees and linear regressions are examples of interpretable models~\cite{Fernandez2019}.

Conversely, opaque models have internal mechanics that are a mystery because humans cannot examine how these intelligent systems function. Even if one could look inside these models, their internal mechanisms would be so complex that making sense of their predictions would be impossible. XAI methods that extract insights about feature importance in black-box models are called model-agnostic.

Explainable machine learning approaches can also be classified into two categories based on their scope: local and global interpretability. Global interpretability corresponds to the overall set of features that contribute to the predictions of a general predictive system. They enable a general comprehension and understanding of the predictive system~\cite{Miller2019}. 
Alternatively, local interpretability is concerned with generating interpretations for a specific local data point rather than providing the overall interpretations of the predictive system. It corresponds to generating interpretations in a specific area of the input space. The decision surface of the model becomes smoother as the input space is restricted. Local interpretability is often achieved through local example-based techniques or local surrogates, which simulate a limited region surrounding an example~\cite{Ribeiro2016,Guidotti2019survey,Fernandez2019,Moreira2021DSS}.

In opaque models, explainability may be achieved through various algorithms, particularly in explanations that rely on feature attribution~\cite{shrikumar2017learning, Sundararajan2017}. The feature attribution-based explanation is a local approach that can provide a score or ranking over features, conveying each feature's (relative) importance to the model's output. LIME~\cite{Ribeiro2016} is one of the most representative attribution-based explainable algorithms in the literature that approximates the local decision boundary to a data point. More specifically, LIME perturbs a sample around the input vector near a local decision boundary~\cite{Elshawi2019}. Each feature is assigned a weight based on a similarity function that compares the distances between the original instance prediction and the sampled locations in the decision boundary's predictions\cite{Ribeiro2016}. Another important feature-attribution algorithm is SHAP which distributes the values of the features in a game theoretic approach. SHAP estimates Shapley values from coalitional game theory to properly share the gain among players so that the contributions of players are fair~\cite{Lundberg2017}.

\subsection{Related Work}

Some surveys and benchmarks on counterfactuals have been recently proposed in the literature. Although some of these benchmarks cover many algorithms, none deeply discuss the counterfactual generation process relative to the algorithm's properties and underlying predictive model. A unique contribution of this work that distinguishes itself from the current benchmark studies is extending the evaluation process to a counterfactual inspection analysis of the counterfactual generation process by analyzing decision paths. Another unique contribution of this work is in alerting to the biased conclusions that one may arrive at if blindly benchmarking counterfactual algorithms without considering their properties (for instance, plausibility). 

The following studies present surveys/benchmarks on counterfactual algorithms related to this work. We recommend that the reader look at these studies to gain a different perspective on counterfactual evaluation.

Artelt \& Hammer\cite{Artelt2019} provide a detailed survey and description of several model-specific counterfactual algorithms based on their mathematical formalisms and how these different counterfactuals could be generated from different underlying predictive models (e.g., decision trees, support vector machines, etc.). However, Artelt \& Hammer\cite{Artelt2019} do not provide any taxonomy for the surveyed counterfactual algorithms, nor do they provide any benchmarking of the algorithms or experimentation.

Verma et al.~\cite{Verma2020} collected a set of $29$ explainable counterfactual algorithms, both model agnostic and model specific, and classified them into different themes. These themes consist of several properties that the authors considered relevant for generating counterfactual algorithms. These themes include (1) the type of model access (either if the counterfactual algorithm requires access to the entire internal mechanics of the predictive model, only to the model's gradients, or only to the prediction of the model); (2) model agnostic (the domain which the counterfactual algorithm can operate on, e.g., model specific); (3) optimization amortization (whether the optimization function of the algorithm can generate single or diverse counterfactuals). The authors provide a comprehensive list of open research challenges and do not provide a detailed description or benchmarking of the listed counterfactual algorithms.

Stepin et al.~\cite{Stepin2021} present a systematic literature review on counterfactuals and contrastive explanation methods for XAI. According to the authors, counterfactual explanations are very similar to contrastive explanations in that they both compute the minimum set of features that need to be changed to get a \textit{constrastive} (different) predictive result. A counterfactual is a contrastive explanation where it is possible to imagine hypothetical scenarios in which a particular condition must be met to achieve a specific outcome. Under this point of view, counterfactuals can be used to explain consequences in contrastive (imagined) scenarios. In their literature review,  Stepin et al.\cite{Stepin2021} propose a taxonomy to classify the different approaches in the literature that use contrastive and counterfactual explanations. However, little is discussed in terms of the formal definitions of the algorithm, and no benchmark was conducted in their work.

Karimi et al.~\cite{Karimi2021survey} also present a systematic literature review with a detailed taxonomy focusing on algorithmic recourse, which consists of methods that can compute the set of actions that can reverse an unfavorable prediction across a range of counterfactual scenarios~\cite{Venkatasubramanian2020}. Karimi et al.~\cite{Karimi2021survey} present several counterfactual properties that overlap this study. However, they do not make a detailed analysis of the algorithms surveyed or evaluate them quantitatively.

Keane \& Smith~\cite{Keane2020} define the notion of a \textit{good counterfactual}, which aligns with the findings of our study: that relying uniquely on measures such as proximity, sparsity, or plausibility is not enough to generate good counterfactuals. By good counterfactuals, Keane \& Smith refer to meaningful counterfactuals to a human user. It is also related to Karimi et al. concept of \textit{recourse} since a counterfactual to be actionable will need to be meaningful and interpretable by a user. In their study, Keane \& Smith~\cite{Keane2020} propose to generate counterfactuals using case-based reasoning to find patterns of good counterfactuals in the data. Although the authors' conclusions partially align with the findings in this study, they did not conduct any rigorous survey of existing counterfactual algorithms, make any benchmark, or attempt to evaluate existing algorithms.

Bodria et al.~\cite{Bodria2021} make an extensive benchmark for several explanation methods (ranging from feature attribution to rule-based methods and counterfactuals). Contrary to this paper, the authors did not make a deep and extensive discussion on counterfactual explanation algorithms and did not analyze the counterfactual generation process (which we present in this study). Also, the predictive models used were restricted to a linear white-box model (a logistic regression) and two tree-based algorithms (XGBoost and Catboost). In our study, we investigate the impact of the type of predictive model used by benchmarking several counterfactuals with a decision tree (white box), a random forest (grey box), and a neural network (black box).

The closest works in the literature that relate to our study are from Mazzine \& Martens~\cite{Mazzine2021survey}, Pawelczyk et al.~\cite{Pawelczyk2021}, and more recently, Guidotti~\cite{Guidotti2022}. These studies make an extensive benchmark and cover more counterfactual algorithms than this study. However, what distinguishes this work from theirs is (1) the analysis of the impact of different machine learning algorithms in instance-based counterfactual algorithms~\cite{Chou2022fusion}, (2) an extensive counterfactual inspection analysis using decision-trees to guide the counterfactual generation process, and (3) a deep discussion on the impact of the different counterfactuals loss functions on the generation of explanations. Although this work covers a smaller amount of counterfactuals (by surveying the instance-based category of counterfactuals algorithms from Chou et al.~\cite{Chou2022fusion}), we were able to do a unique benchmark which enabled us to investigate the counterfactual generation process in a different perspective from other studies in the literature. Since our analysis is different, the findings we retrieved from this study are also singular, and only Keane \& Smith~\cite{Keane2020} partially concluded some of the findings that we put forward in this work with their case-based reasoning approach to counterfactual explanations. 

\section{Counterfactuals in XAI}

Unlike attribute-based algorithms that assign a significance score to each input feature, counterfactuals generate examples grounded on the underlying predictive model with the minimum number of changes relative to the input vector~\cite{Molnar2020book}. In the scientific community, counterfactuals are valued for their ability to provide humans with causal and understandable explanations since they promote mental representations of actual and alternate events~\cite{Miller2019,Holzinger19}. A counterfactual approach for explanations is one of the most promising methods to achieve responsible AI since it can potentially satisfy GDPR’s policy requirements for explainability~\cite{Holzinger21,Watcher2018,Miller2019}. 

Although counterfactuals provide a different mechanism for generating explanations compared to feature attribution methods, some works in the literature have tried to propose a unification method for feature attribution and counterfactual explanations~\cite{Mothilal2021}. From a cognitive science/psychology perspective, some works emphasize the importance of counterfactual thinking in social scenarios. For instance, Pereira and Santos~\cite{Pereira2019}, used counterfactuals to understand how individuals that used counterfactual reasoning could improve cooperation in populations. Their models used evolutionary game theory and found that a small presence of individuals using counterfactual thinking was enough to nudge an entire population towards highly cooperative standards. In Pereira and Barata~\cite{Pereira2020}, the authors argue that counterfactuals are important ingredients to building machines with adequate moral capacity. 

\subsection{Generation of Counterfactuals}\label{ref-pro} 
Counterfactual instances can be found by iterative perturbing the input features of the test instance until the desired prediction is obtained~\cite{Lewis1973counterfactual}. It measures the smallest change between a data instance and a counterfactual instance~\cite{Watcher2018}. This notion is described in Equation~\ref{eq:initialCF}, where $d(.,.)$ is a measurement for determining the smallest distance between a data point $x$ and the counterfactual $x'$ and parameter $\lambda$ balances the distance in the prediction against the distance in feature values~\citep{Molnar2020book}. The higher the value of $\lambda$, the closer the counterfactual candidate, $x'$, is to the desired outcome, $y'$.

  \begin{equation}
            \begin{split}
                \mathcal{L}(x, x', y', \lambda) = \lambda \left( f(x') - y'\right)^2 + d(x,x') \\
                arg~\underset{x'}{min}~\underset{\lambda}{max} \mathcal{L}(x, x', y', \lambda)~~~~~~~~~~~~~~~~~
            \end{split}
            \label{eq:initialCF}
        \end{equation}

Several distance functions have been proposed in the literature: the L$_1$-norm, the L$_2$-norm, and the L$_\infty$-norm.

The L$_1$-norm (also known as Manhattan distance) is the most explored distance function in the literature of counterfactuals in XAI~\citep{Mothilal2020DiCE,Russell2019,Grath2018,Pawelczyk2020CCHVAE}. It was initially proposed in the work of Watcher~\cite{Watcher2018} was the first to propose this norm in a loss function to find the counterfactual with the minimum instance from its original input.
 
The L$_2$-norm (also known as Euclidean distance) calculates the shortest distance between two points but does not necessarily yield sparse solutions due to its circular shape, and it is more sensitive to outliers~\cite{Molnar2020book}. 

\begin{figure}[!h]
        \resizebox{\columnwidth}{!}{
        \includegraphics{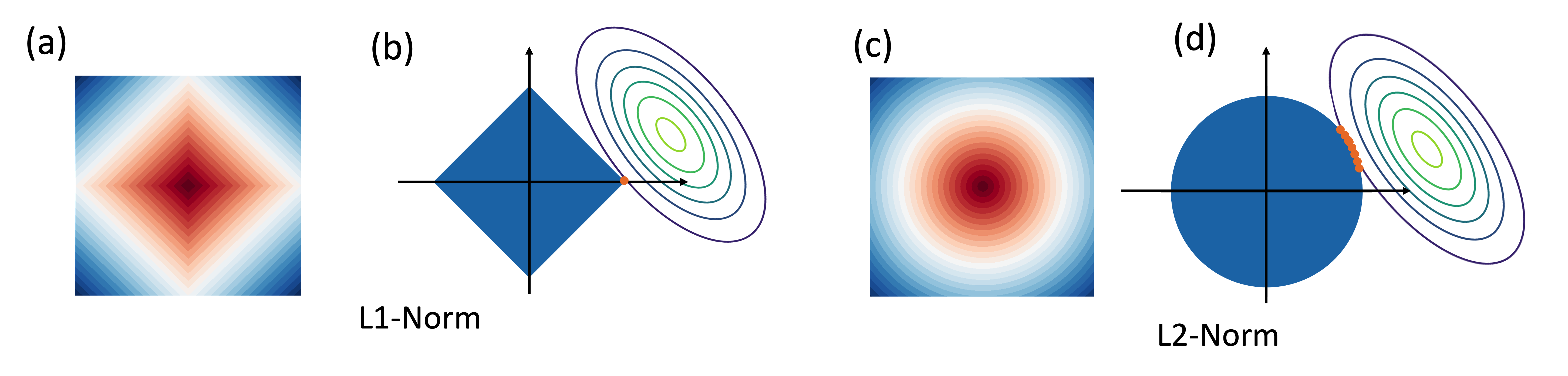}
        }
        \caption{Impact of different norms in sparsity. L$_1$-norm promotes sparseness because of its diamond shape function. The intersection of a vector with one of the function's corners will lead to a sparse result, whereas in the figure, only the $x$ coordinate will have a value different from $0$. This is not true for the L$_2$-norm due to its circular shape.}
        \label{fig:norm_and_sparsity}
\end{figure}

Figure~\ref{fig:norm_and_sparsity} shows how different types of norms impact the sparseness of the data. L$_1$-norm promotes sparseness because of its diamond shape function. The intersection of a vector with one of the function's corners will lead to a sparse result. In Figure~\ref{fig:norm_and_sparsity}, only the $x$ coordinate will have a value different from $0$. This is not true for the L$_2$-norm due to its circular shape. However, the circular shape of the $L_2$-norm promotes a differentiable function, while in the $L_1$-norm, this operation becomes harder. In counterfactual search, authors often must deal with trade-offs between the data sparseness and the functions' differentiability and apply appropriate optimization methods to generate counterfactuals.

Karimi et al.~\cite{Karimi2020mace} present the first study to investigate the L$\infty$-norm for counterfactuals in XAI. In this formulation, the cost of the most significant features is penalized for limiting the maximum change across features between a given initial instance $x$ and a given counterfactual candidate $x'$. This process reduces the L$\infty$-norm and leads to less sparse solutions than other norms. For more information about distance functions in counterfactual explanations, please refer to the literature review of Chou et al.~\cite{Chou2022fusion}.

\subsection{Properties of Counterfactuals}

\begin{figure}[!ht]
    \centering
    \includegraphics[scale=0.4]{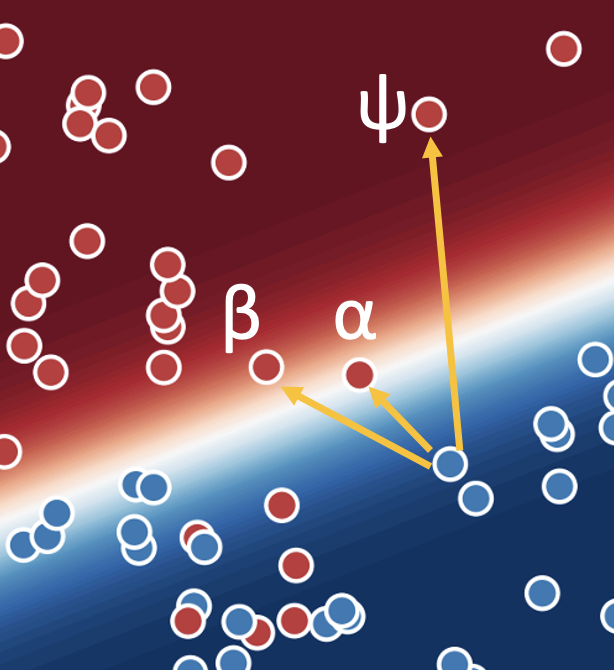}
    \caption{The different counterfactual candidates for a data instance $x$. According to Watcher~\cite{Watcher2018}, counterfactual $\alpha$ is the best candidate because it has the shortest Euclidean distance to $x$. Other researchers believe that counterfactual instance $\Psi$ is the best option because it gives a feasible path from $x$ to $\Psi$~\cite{Poyiadzi2020}. Counterfactual $\beta$ is another candidate of poor quality because it lies in a less defined region of the decision boundary.}
    \label{fig:counterfactual_general}
\end{figure}

Several studies consider a set of properties to assess the quality of a generated counterfactual~\cite{Keane2020,Stepin2021}. They can be summarised as follows:

\begin{itemize}
    \item \textbf{Proximity.} This property states that a good counterfactual must have the smallest distance to its original feature vector. A small distance translates into fewer features changed, which increases the human interpretability of the explanation~\citep{Verma2020}.
    
    \item \textbf{Plausibility.} This property is analogous to \textit{Actionability} and \textit{Reasonability}~\cite{Keane2020, Verma2020,Prosperi2020,Poyiadzi2020,Rawal2020}. Counterfactuals that are plausible need to be valid, and the search process should yield logically plausible outcomes. As a result, \textit{immutable} features should never be changed (such as religion and gender).
    
    \item \textbf{Sparsity.} This property consists of finding the minimal feature set that must be modified to obtain a counterfactual~\cite{Keane2020}. Ideally, a counterfactual should be sparse to promote user understandability: the fewer features that need to be changed to generate a counterfactual, the more understandable it becomes for the user~\cite{Mothilal2020DiCE}.
    
    \item \textbf{Diversity.} Counterfactual explanations can suggest changes in features that are not easily understandable to certain user groups~\citep{Russell2019,Karimi2021survey}. Diversity overcomes this problem by generating different counterfactuals while preserving low proximity and sparsity~\citep{Mothilal2020DiCE}.
    
    \item \textbf{Feasibility.} Counterfactual explanations solely based on the minimum feature change can suggest modifications in features that are not feasible or practical for a user to implement and to achieve the desired predictive outcome~\cite{Poyiadzi2020}. Figure~\ref{fig:counterfactual_general} shows an overall example where the generated counterfactual $alpha$ corresponds to the shortest distance to the original input vector "$x$" but also falls within the decision boundary, which has predictions with the highest levels of uncertainty. As a result, counterfactual explanations may be biased. A more preferred counterfactual would be $\Psi$ since it falls in a distinct region of the decision space while still preserving the shortest path to the original vector $x$.
    
\end{itemize}

Note that the plausibility and actionability definitions in this paper align with a functional-grounded scope. Defining actionability from a user perspective can be challenging due to its subjective nature and the varying needs of different users.  This is because what is considered actionable can vary significantly depending on the user's context, goals, and expertise (see~\cite{klein_jalaeian_hoffman_mueller_2021,Kirfel2021if}).

\subsection{Model-Agnostic Counterfactual Generation Methods in XAI}

In their prior work,~\cite{Chou2022fusion} showed that many counterfactual algorithms shared similar theoretical backgrounds. The authors analyzed 23 model-agnostic XAI counterfactual techniques and categorized them into seven groups, each reflecting their underlying "master theoretical algorithm"~\cite{Domingos2017}. We provide a summary of the seven main categories. Note that no single classification can perfectly capture the complexity of the rapidly evolving field of model-agnostic counterfactuals. This taxonomy was chosen because it provides a clear and comprehensive framework for model-agnostic XAI counterfactuals. For more details, the reader can refer to the work of Chou et al.~\cite{Chou2022fusion}.

\begin{itemize}
    \item  \textbf{Instance-Centric.} These approaches mainly focus on developing loss functions that generate counterfactuals based on the minimum distance to the original feature vector~\cite{Watcher2018, Lewis1973counterfactual}. These methods are more likely to fail the plausibility, feasibility, and diversity properties because they consist of finding the minimum distance for different L$_p$-norms without constraining features~\cite{Grath2018,Looveren2019,Laugel2018GrowingSpheres,Laugel2019,Laugel2020unjustified}.
    
    \item  \textbf{Constraint-centric.} These approaches model loss functions as constraint satisfaction problems~\cite{Karimi2020mace, Russell2019}. The optimization process is guided by constraints, which specify which features should not be changed. Thus, these approaches satisfy different counterfactual properties, such as feasibility, diversity, and plausibility.
    
    \item  \textbf{Genetic-centric.} These approaches use genetic algorithms in loss function optimization to search for counterfactuals. Due to the ability of genetic search to allow cross-overs and mutations, these approaches often fulfill properties such as proximity and diversity~\cite{Guidotti2019survey,Sharma2020,Dandl2020}.
    
    \item  \textbf{Regression-centric.} These approaches operate similarly to LIME~\cite{Ribeiro2016}. Their loss function is based on a linear regression model, and the weights of this model are presented as explanations. Counterfactuals based on these approaches have difficulties satisfying several properties such as plausibility and diversity~\cite{White2020,Ramon2020}.
    
    \item  \textbf{Game Theory-centric.} These approaches operate similarly to SHAP~\cite{Lundberg2017} and generate explanations using Shapley values. It consists mainly of algorithms that extend the SHAP algorithm to consider counterfactuals~\cite{Ramon2020,Rathi2019}. The counterfactuals derived from these approaches fail to satisfy most properties, such as plausibility and diversity.
    
    \item  \textbf{Case-Based Reasoning Centric.} These approaches model loss functions inspired by the cognitive science case-based reasoning paradigm, which portrays the reasoning process as essentially memory-based~\cite{Keane2020}. These methods frequently create new counterfactuals by retrieving previously generated counterfactuals. These approaches can easily satisfy different properties, such as diversity, plausibility, and feasibility.
    
    \item  \textbf{Probabilistic-Centric.} The counterfactual generation problem is modeled as a probabilistic problem in this category. Random walks, Markov sampling, variational autoencoders, and probabilistic graphical models are frequently used in these approaches to learning efficient data codings~\cite{Ghazimatin20,Pawelczyk2020CCHVAE,Lucic2020,Guidotti2020,Downs2020crud, Karimi2020nips, Rawal2020, Barocas2020}. Probabilistic approaches have the potential to meet the causality framework suggested by Pearl~\cite{Pearl2009} and generate less biased counterfactuals.\\
\end{itemize}

\subsection{Instance-Centric Counterfactual Algorithms}
In this study, we evaluate the performance of several machine learning models using instance-centric approaches. We focus on this approach because of its popularity and simplicity. Most importantly, we consider that instance-centric approaches constitute the basis of most counterfactual algorithms in the literature. In this study, we will explore four instance-centric counterfactual explainable algorithms: WatcherCF~\cite{Watcher2018}, Prototype~\cite{Looveren2019}, GrowingSpheresCF~\cite{Laugel2018GrowingSpheres}. Table~\ref{tab:algo_summary} summarizes the features of these instance-centric counterfactual algorithms in terms of several properties. This table is the result from the previous survey of Chou et al.\cite{Chou2022fusion} and will serve as a basis for this benchmark. For a full description of these algorithms, please refer to Chou et al.~\cite{Chou2022fusion}.

\begin{itemize}
    \item \textbf{WatcherCF~\cite{Watcher2018}.}  The goal is to find a counterfactual $x'$ with the minimum distance to the original data point $x_i$. Several different norms can be used as distance functions. However, the authors propose using the L$_1$-norm since it promotes sparsity.
    
    \item \textbf{Prototype Counterfactuals~\cite{Looveren2019}.} This algorithm extends the contrastive explanation method (CEM)~\cite{Dhurandhar2018}, which generates counterfactuals in terms of Pertinent Positives (PP) and Pertinent Negatives (PN). Using this method, one can determine what features are minimally and sufficiently required to predict the same class as the original instance (the PFs). One can also use this method to identify which features should be minimized and necessarily absent from the instances (the PNs). Prototype extends this method by adding a prototype loss term in the objective result to generate more interpretable counterfactuals. 
    
    \item \textbf{GrowingSpheresCF~\cite{Laugel2018GrowingSpheres,Laugel2019,Laugel2020unjustified}.} GrowingSpheresCf Counterfactual Explanations address the problem of determining the minimal changes to alter a prediction by proposing an inverse classification approach. The authors present the \textit{Growing Spheres} algorithm, which consists of identifying a close neighbor classified differently through the specification of sparsity constraints that define the notion of \textit{closeness}. 
    
    \item \textbf{DiCE~\cite{Mothilal2020DiCE}.} Diverse Counterfactual Explanations generate diverse counterfactual explanations for the same data instance $x$, allowing the user to choose more understandable and interpretable counterfactuals. Diversity is formalized as a determinant point process, which is based on the determinant of the matrix containing information about the distances between a counterfactual candidate instance and the data instance to be explained.
\end{itemize}

\begin{table}[!h]
\caption{Classification of Instance-centric model-agnostic algorithms as proposed by Chou et al. \cite{Chou2022fusion}.}
\label{tab:algo_summary}
\resizebox{\columnwidth}{!} {
\begin{tabular}{cccc|c|c|c|c|c|c|}
\cline{5-10}
& & & & 
  \multicolumn{6}{c|}{\textbf{Properties}} \\ \hline
  \multicolumn{1}{|c|}{\textbf{Algorithms}} &
  \multicolumn{1}{c|}{\textbf{Ref.}} &
  \multicolumn{1}{c|}{\textbf{Applications}} &
    \multicolumn{1}{c|}{\textbf{Code?}} &
  \textbf{Proximity} &
  \textbf{Plausibility} &
  \textbf{Sparsity} &
  \textbf{Diversity} &
  \textbf{Feasibility} &
  \textbf{Optimization}  \\ \hline
  
  \multicolumn{1}{|l|}{WatcherCF} &
  \multicolumn{1}{c|}{\citep{Watcher2018}} &
  \multicolumn{1}{c|}{\begin{tabular}[c]{@{}c@{}}C\\ {[}Tab / Img{]}\end{tabular}} &
  \begin{tabular}[c]{@{}c@{}}Yes \citep{Alibi}\\ {[}Algo: CF{]}\end{tabular} &
  \begin{tabular}[c]{@{}c@{}}\greencheck\\ {[}L$_1$-norm{]}\end{tabular} &
  \crosscheck &
  \greencheck &
  \crosscheck &
  \crosscheck &
  Gradient Descent \\ \cline{1-10} 

  \multicolumn{1}{|l|}{\begin{tabular}[c]{@{}l@{}}Prototype \\ Counterfactuals\end{tabular}} &
  \multicolumn{1}{c|}{\citep{Looveren2019}} &
  \multicolumn{1}{c|}{\begin{tabular}[c]{@{}c@{}}C\\ {[}Tab / Img{]}\end{tabular}} &
  \begin{tabular}[c]{@{}c@{}}Yes \citep{Alibi}\\ {[}Algo: CFProto{]}\end{tabular} &
  \begin{tabular}[c]{@{}c@{}}\greencheck\\ {[}L$_1$/L$_2$-norm{]}\end{tabular} &
  \greencheck &
  \begin{tabular}[c]{@{}c@{}}\greencheck\\ {[}kd-trees / auto-encoders{]}\end{tabular} &
  \crosscheck &
  \crosscheck &
  FISTA \\ \cline{1-10} 
  
  \multicolumn{1}{|l|}{Growing Spheres} &
  \multicolumn{1}{c|}{\citep{Laugel2018GrowingSpheres,Laugel2020unjustified,Laugel2019}} &
  \multicolumn{1}{c|}{\begin{tabular}[c]{@{}c@{}}C\\ {[}Tab / Txt / Img{]}\end{tabular}} &
  Yes \citep{GrowingSpheres} &
  \begin{tabular}[c]{@{}c@{}}\greencheck\\ {[}L$_0$-norm{]}\end{tabular} &
  \crosscheck &
  \greencheck &
  \crosscheck &
  \crosscheck &
  Growing Spheres  \\ \cline{1-10} 
  
  \multicolumn{1}{|l|}{DICE} &
  \multicolumn{1}{c|}{\citep{Mothilal2020DiCE}} &
  \multicolumn{1}{c|}{\begin{tabular}[c]{@{}c@{}}C\\ {[}Tab{]}\end{tabular}} &
  Yes \citep{DICE} &
  \begin{tabular}[c]{@{}c@{}}\greencheck\\ {[}L$_1$-norm{]}\end{tabular} &
  \greencheck &
  \greencheck &
  \greencheck &
  \greencheck &
  Gradient Descent  \\ \hline
\end{tabular}
}
\end{table}

\section{A Benchmark Evaluation of XAI Counterfactual Algorithms}

This study proposes a benchmark framework to assess XAI counterfactual algorithms using state-of-the-art quantitative metrics and a counterfactual inspection analysis. The following sections present the proposed experimental design, the datasets used, and the quantitative metrics we applied in our benchmark.

\subsection{Experimental design}

We designed our experiments in terms of four major phases: (1) dataset selection and pre-processing; (2) model training and evaluation; (3) counterfactual explanation generation; (4) counterfactual explanation quantitative analysis; and (5) a counterfactual inspection analysis.

\begin{figure}[!h]
    \centering
    \resizebox{\columnwidth}{!} 
    {
    \includegraphics[scale=0.4]{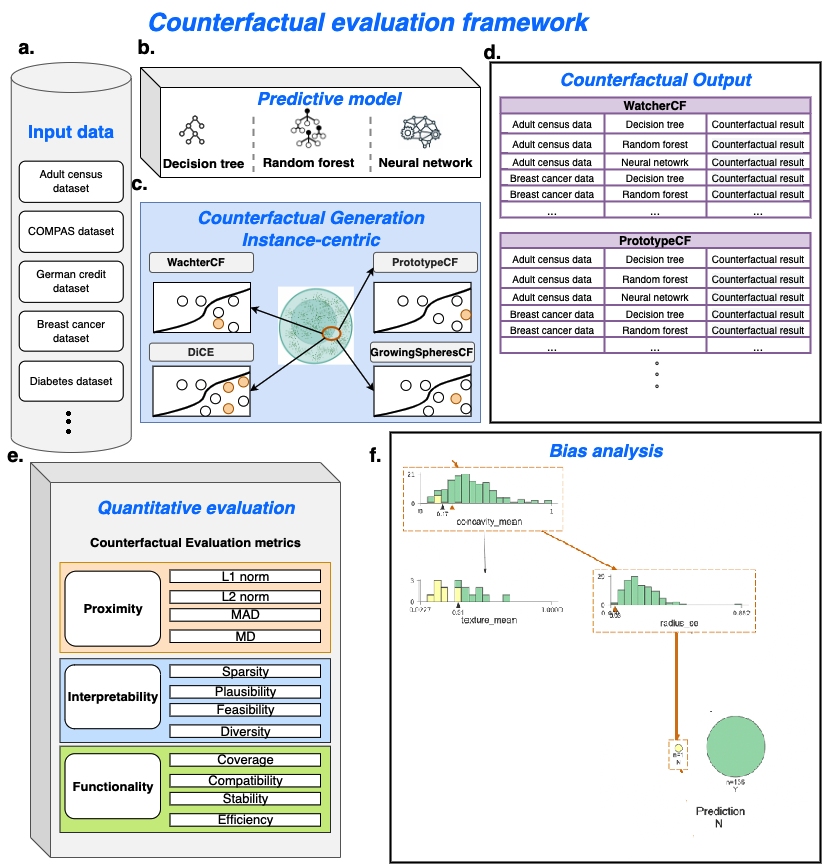}
    }
    \caption{Experimental design. Panel \textbf{a.} shows the five datasets used. Panel \textbf{b.} shows the three types of predictive models trained for each dataset. Panels \textbf{c.} and \textbf{d.} show the explainable counterfactual algorithms used and the corresponding results (the counterfactual explanations). Panel \textbf{e.} shows the metrics used to evaluate the generated counterfactuals (quantitative analysis). Panel \textbf{f.} shows the counterfactual inspection analysis used to assess the quality of the generated counterfactual explanations 
    (counterfactual inspection).
    }
    \label{fig:experimental_design}
\end{figure}

These five experimental phases were designed to analyze two main aspects. First, to investigate the impact of different types of predictive models in the counterfactual generation process. Second, to explore mechanisms allowing a decision-maker to inspect and assess the generated counterfactuals' quality: whether the explanations are biased due to biased decision points learned by the predictive model or if the generated explanation is meaningful to a human decision-maker. This last point is one of this study's significant and singular contributions and what differentiates this benchmark from other proposed benchmarks in the literature. Figure~\ref{fig:experimental_design} presents the overall experimental design for our benchmark study. The following sections provide details of each of these five phases.

\subsection{Datasets}
 We applied these models to 25 tabular datasets presented in \citet{grinsztajn2022tree}. 
 Our benchmark comprises 17 datasets with solely numerical features and eight with a mix of numerical and categorical features. Three datasets were pre-processed by the authors, where we used one hot encoding in their categorical features (adult, credit, and compass), and the remaining five datasets used the pre-processing of~\citet{grinsztajn2022tree}, which combine binary and ordinal encoding. By having different datasets with different feature encoding mechanisms, we can analyze how the different counterfactual algorithms are affected by more sparse inputs (one hot encoding) or less sparse vectors. Table~\ref{tab:datainfo_table} presents an overall data description of the datasets used. Note that we separated numerical datasets from mixed datasets because WachterCF and GrowingSpheresCF do not work with categorical data,

\begin{table}[!h]
\caption{Data information: The table lists the information for three categorical data and two numerical data. To accurately quantify the feature, we used one-hot encoding to transform the categorical data into a format that could be fed into the machine learning model for prediction. \textcolor{blue}{All datasets have binary classification labels.}}
\label{tab:datainfo_table}
\resizebox{\columnwidth}{!}{
\begin{tabular}{|l|l|l|l|l|l|l|l|}
\hline
\textbf{Dataset} &
  \textbf{Type} &
  \textbf{\begin{tabular}[c]{@{}l@{}}Sample \\ Size\end{tabular}} &
  \textbf{\begin{tabular}[c]{@{}l@{}}Total \\ Features\end{tabular}} &
  \textbf{\begin{tabular}[c]{@{}l@{}}Numeric \\ Features\end{tabular}} &
  \textbf{\begin{tabular}[c]{@{}l@{}}Categorical \\ Features\end{tabular}} &
  \textbf{\begin{tabular}[c]{@{}l@{}}Type of \\ Enconding\end{tabular}} &
  \textbf{\begin{tabular}[c]{@{}l@{}}Encoded \\ Features\end{tabular}} \\ \hline
Electricity       & mixed & 38474  & 8  & 7  & 1  & OrdEnc & 1   \\ \hline
Eye Movements     & mixed & 7608   & 23 & 20 & 3  & BinEnc & 3   \\ \hline
Covertype         & mixed & 423680 & 54 & 10 & 44 & BinEnc & 44  \\ \hline
Albert            & mixed & 58252  & 31 & 21 & 10 & OrdEnc & 10  \\ \hline
Road safety       & mixed & 111762 & 32 & 29 & 3  & BinEnc & 3   \\ \hline
Adult             & mixed & 32651  & 12 & 4  & 8  & OHE    & 103 \\ \hline
German            & mixed & 1000   & 20 & 5  & 15 & OHE    & 65  \\ \hline
COMPAS            & mixed & 7214   & 11 & 4  & 7  & OHE    & 23  \\ \hline
California        & num   & 20634  & 8  & 8  & 0  & -      & 8   \\ \hline
Credit            & num   & 16714  & 10 & 10 & 0  & -      & 10  \\ \hline
Heloc             & num   & 13488  & 22 & 22 & 0  & -      & 22  \\ \hline
Jannis            & num   & 57580  & 54 & 54 & 0  & -      & 54  \\ \hline
Diabetes130US     & num   & 71090  & 7  & 7  & 0  & -      & 7   \\ \hline
Eye Movements     & num   & 7608   & 20 & 20 & 0  & -      & 20  \\ \hline
Higgs             & num   & 940160 & 24 & 24 & 0  & -      & 24  \\ \hline
Default of Credit & num   & 13272  & 20 & 20 & 0  & -      & 20  \\ \hline
MiniBooNE         & num   & 72998  & 50 & 50 & 0  & -      & 50  \\ \hline
Bank Marketing    & num   & 10578  & 7  & 7  & 0  & -      & 7   \\ \hline
Magic Telescope   & num   & 13376  & 10 & 10 & 0  & -      & 10  \\ \hline
House 16H         & num   & 13488  & 16 & 16 & 0  & -      & 16  \\ \hline
Pol               & num   & 10082  & 26 & 26 & 0  & -      & 26  \\ \hline
Covertype         & num   & 566602 & 10 & 10 & 0  & -      & 10  \\ \hline
Electricity       & num   & 38474  & 7  & 7  & 0  & -      & 7   \\ \hline
Pima Diabetes     & num   & 768    & 9  & 9  & 0  & -      & 9   \\ \hline
Breast Cancer     & num   & 569    & 30 & 30 & 0  & -      & 30  \\ \hline
\end{tabular}
}
\end{table}


\subsection{Predictive Models}

In this benchmark, we tested different types of machine learning models to understand how different predictive models impact the counterfactual generation process. 

\begin{itemize}
    \item \textbf{White box model.} It is a model with clear underlying logic and programming processes, making its decision-making process inherently interpretable~\cite{Pintelas2020}. In this study, we selected a decision tree as an example of a white box model for our experiments.
    
    \item \textbf{Grey box model.} It is a model that combines the capabilities of white-box models with black-box models~\citep{Bohlin2006}, leading to models that are both accurate and semi-interpretable. In this study, we selected a random forest as an example of a grey box model for our experiments. The random forest leans on an ensemble of trees to make a prediction. Although the trees are white box models, the ensemble nature of the model makes it very hard for a human to understand how the prediction was computed~\cite{Wanner2020}.
    
    \item\textbf{Black-box model.} It is an ML model whose inner workings are so complex that they become difficult for a human user to understand. In this study, we selected a deep neural network as an example of a black box model for our experiments.
\end{itemize}

After processing the data, we divided the datasets into two groups: the training set (80\%) and the test set (20\%). We used the training set to fit the data to the different machine learning models and the test set to evaluate the ML model and generate counterfactual explanations. The ML models were trained to have similar performances in terms of accuracy, precision, recall, and F1-score (see Appendix, Table~\ref{tab:classification_metrics} for detailed results on each ML model's performance for each dataset).


\subsection{Evaluation Metrics}

In this study, we evaluated the generated counterfactual explanations according to three groups of metrics: (1) \textit{proximity metrics}, which are primarily focused on measuring the distance between counterfactuals; (2) \textit{interpretability metrics}, which consist in determining the smallest number of features to be changed in the counterfactual, this way promoting user understandability; and (3) \textit{functionality metrics}, which consist in measuring the performance of the counterfactual generation process.

The following evaluation metrics were collected from multiple studies from the XAI counterfactual literature~\cite{Mazzine2021survey,Mothilal2020DiCE,Looveren2019,Laugel2018GrowingSpheres}.\\

\noindent
\textbf{Proximity Metrics.} It consists of metrics for determining the distance between the initial instance and the generated counterfactual from the instance~\cite{Russell2019,Mothilal2020DiCE,Laugel2018GrowingSpheres}. It considers the variation of each feature changed. In this study, we considered the following proximity metrics:
\begin{itemize}
    \item \textbf{L$_1$-Norm.} Measures the absolute difference between a data instance $x$ and the counterfactual candidate $x'$. 
    \begin{equation}
        L_1Norm(x, x') = \|x - x'\|_1 = \sum_{j}^p |x_j - x'_j|
    \end{equation}
    
    \item \textbf{L$_2$-Norm.} Measures the square root of the sum of the squared vector values between a data instance $x$ and the counterfactual candidate $x'$.
    
    \begin{equation}
           L_2Norm(x, x') = \|x - x'\|_2 = \sum_{j}^p \sqrt{ x_j^2 - x_j'^2}.
    \end{equation}
    
    \item \textbf{Inverse of Median Absolute Derivation (IMAD).} Given a  Consists of the $L_p$-norm normalized by the inverse of the median absolute deviation of feature $j$ over the dataset is one of the best-performing distance functions because it ensures the sparsity of the counterfactual candidates. This work used the IMAD function as a normalization factor for the L$_1$-norm (Equation~\ref{eq:mad}).
         \begin{equation}
    \begin{split}
         IMAD(x,x') = \sum_{j}^p \frac{|x_j - x'_j|}{MAD_j}, \text{~ where~~~~~~~~~~~~~~~~~~~~~~~~~~~~~~~~~~~~~~~~}\\
         MAD_j = median_{ i\in	\{1,\dots,n\}}  \left|x_{i,j} - median_{l\in \{1,\dots,n\}}(x_l,j)\right|
    \end{split}
\label{eq:mad} 
\end{equation}
The median absolute deviation is calculated as the median of the absolute deviations from the median of the feature values. By normalizing each feature difference by its median absolute deviation, the IMAD function accounts for the variability in each feature, promoting sparsity in the generated counterfactuals. This normalization helps identify the most significant changes needed to create a counterfactual, ensuring that only a few features are altered significantly.

    \item \textbf{Mahalanobis Distance (MD).} Given a data instance array $x$ and a counterfactual candidate array $x'$, their Mahalanobis distance is defined by 
    \begin{equation}
        MD(x,x') = \sum_j^p \sqrt{ (x_j - x_j') V^{-1} (x_j - x_j')^T},
    \end{equation}
    where $V$ is the covariance matrix. This distance function is often applied to discover multidimensional outliers and to indicate feature correlation~\cite{Mclachlan1999mahalanobis}.\\
\end{itemize}

\noindent
\textbf{Interpretability Metrics.} It refers to metrics that indicate how interpretable an algorithm is. Interpretability metrics favour counterfactual explanations with the fewest feature changes. In this study, we considered the following metrics:
\begin{itemize}
    \item \textbf{Sparsity.} Measures the number of features that changed from a data instance array $x$ and a counterfactual candidate $x'$.
    
    \begin{equation}
        Spa(x,x') =  \sum_j^p \mathds{1}_{x_j \neq x'_j}  
    \end{equation}

    \item \textbf{Sparsity Rate.} Measures the number of features that changed from a data instance array $x$ and a counterfactual candidate $x'$ divided by the array's total number of features, $p$.
    
    \begin{equation}
        SpaRate(x,x') = \frac{1}{p} \sum_j^p \mathds{1}_{x_j \neq x'_j}
    \end{equation}

    \item \textbf{Plausibility.} A qualitative measure that checks whether the XAI counterfactual algorithm can generate counterfactuals that do not change sensitive (or immutable) features (e.g. gender or race).
    
    \item \textbf{Feasibility.} A qualitative measure of whether the 
    XAI algorithm can generate counterfactuals that suggest feature changes that are useful (and feasible) for the decision-maker to take action. Often, feasibility is related to plausibility: feasibility implies that sensitive features are not changed during the counterfactual generation process. Feasibility is also related to actionability~\cite{Karimi2021survey}.
    
    \item \textbf{Diversity.} A qualitative metric that checks whether an XAI counterfactual algorithm can generate different counterfactual explanations. Diversity is important to present the user with alternative counterfactual scenarios so the user can choose the ones that are more understandable or feasible in a given decision problem~\cite{Mothilal2020DiCE}. Note that XAI algorithms that ensure diversity will be penalized in terms of stability and vice-versa. \\
\end{itemize}

\noindent
\textbf{Functionality Metrics.} They refer to metrics related to how efficiently an algorithm generates counterfactuals. In this study, we considered the following functionality metrics:
\begin{itemize}
    \item \textbf{Coverage.} XAI counterfactual algorithms often cannot find a counterfactual explanation at the appropriate time. This metric measures how many times a counterfactual explanation was found in each experimental setting, averaged by the number of executions of the algorithm. In this study, we executed each counterfactual XAI algorithm five times for each data test instance~\cite{Looveren2019}.
    
    \item \textbf{Compatibility.} This qualitative metric measures whether an XAI counterfactual algorithm can process numerical and categorical variables.
    
    \item \textbf{Stability.}  It is a metric that assesses an algorithm's ability to provide consistent results across several runs using the same model and input data~\cite{Mazzine2021survey}. 
    If the XAI counterfactual algorithm outputs the same counterfactual explanation for the same input in two consecutive runs, then we consider this metric to be $1$ (stable); otherwise, $0$ (unstable).
    
    \item \textbf{Efficiency.} It measures how many seconds it takes for a counterfactual algorithm to generate an explanation.
\end{itemize}

\subsection{System Specifications}

To ensure the consistency of the experiments, all the programs were conducted on an Apple M1 chip (64-bit) with an 8-core CPU. For the software setting, we conducted Python 3.7.11 and paired TensorFlow version 1 to satisfy the need for the ALIBI\cite{Alibi} packages and TensorFlow 2 with GrowingSpheresCF and DiCE counterfactual algorithms. The open source code can be accessed in \url{https://github.com/LeonChou5311/Counterfactual-benchmark}.

\section{Experimental Results and Analysis}

For each counterfactual algorithm, we randomly selected $20$ instances of the test set (as proposed in Looveren and Klaise~\cite{Looveren2019}) and ran the counterfactual algorithm $5$ times on each data instance. Ultimately, we generated $100$ counterfactual explanations from different counterfactual algorithms for each machine learning model and evaluated the counterfactuals with the proximity, interpretability, and proximity metrics. The following sections present a detailed analysis and discussion of the results obtained for the counterfactual explanations generated for \textit{numerical datasets} (detailed results can be found in Appendix, Table~\ref{tab:num_data}) and for \textit{mixed-datasets} (Appendix, Table~\ref{tab:mixed_data}).

\subsection{Overall Analysis of Counterfactual Explanation Algorithms}

Four counterfactual explanation algorithms (DiCE, GrowingSpheresCF, Prototype, and WatcherCF) have been tested on three different machine learning models (Decision Trees, Random Forests, and Neural Networks) on $17$ \textit{numerical} datasets and eight mixed data datasets. Figure~\ref{fig:proximity_box_plot} presents a summary of the performance and consistency of counterfactual algorithms for different ML models according to the L$_1$ norm. Results are presented on a logarithmic scale, offering insightful interpretations regarding the performance and characteristics of these algorithms across the various datasets and machine learning models. For detailed results at the dataset level, please refer to Tables~\ref{tab:num_data} and~\ref{tab:mixed_data} in the Appendix. The experimental results allow us to draw the following conclusions:

\begin{figure}[!h]
    \includegraphics[scale=0.5]{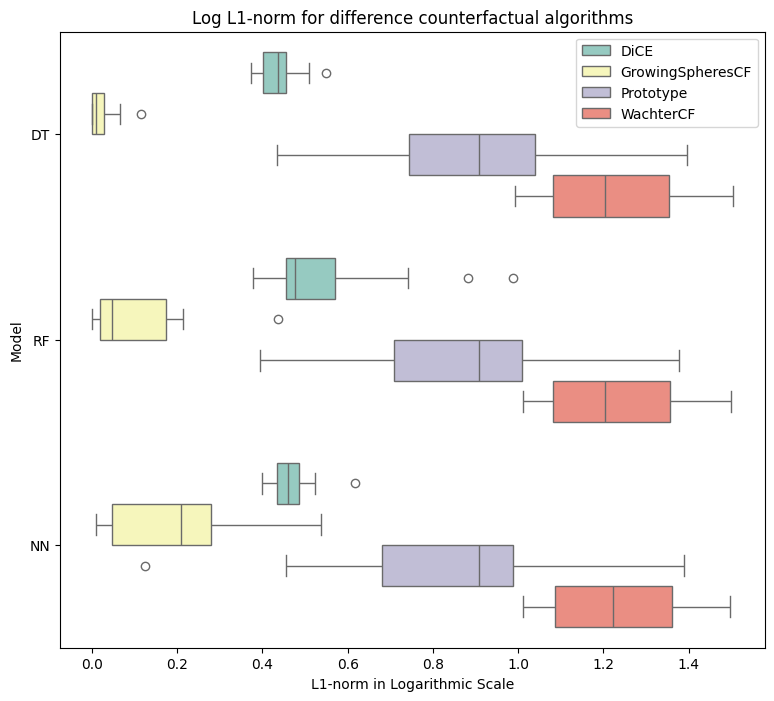}
    \caption{L$_1$ Norm of the generated counterfactual explanations across the different machine learning algorithms: DT corresponds to a Decision Tree, RF to a Random Forest, and NN to a Neural Network.}
    \label{fig:proximity_box_plot}
\end{figure}

\vspace{0.2cm}
\begin{itemize}
    \item \textbf{GrowingSpheresCF generates the best counterfactual explanations in terms of \textit{proximity} and \textit{sparsity}}. 
    
    In Growing Spheres, counterfactuals are generated by defining weighted linear equations of $L_2$-norm and $L_0$-norm between a counterfactual candidate $x'$ and a data instance $x$. The $L_2$-norm minimizes the distance between $x$ and $x'$, while a hyperparameter $\gamma$ ensures that the $L_0$-norm guarantees the generated counterfactual has the least number of features changed, this way promoting sparsity. This way, Growing Spheres optimizes proximity and sparsity in its loss function, resulting in the best counterfactual generator algorithm in our benchmark. However, this algorithm does not ensure \textit{plausibility}, which can lead to biased counterfactual explanations~\cite{Keane2021}. A counterfactual that only satisfies minimum proximity may be biased and incomprehensible to a user if it violates \textit{plausible} changes to sensitive features. For example, \textit{To have diabetes, you will need to go through 50 pregnancies~\cite{Keane2021}}. Equation~\ref{eq:gs} presents the loss function of GrowingSpheresCF.
    
    \begin{equation}
        \begin{split}
            x^* = arg \underset{x'\in X}{min} \{c(x,x')~|~f(x') \neq f(x)\} \\ 
            c(x,x') = \left|\left| x-x' \right|\right|_2 + \gamma\left| \left|x-x'\right|\right|_0
        \end{split}
        \label{eq:gs}
    \end{equation}   
    
    The hyperparameter $\gamma$ plays an important role in the optimization process and in generating sparse counterfactuals. Figure~\ref{fig:GS_analysis} analyses the impact of $\gamma$ in the Growing Spheres loss function. With $\gamma=1$, the loss function combines the $L_2$-norm with the $L_0$-norm. Although the $L_0$-norm is non-differentiable, the $L_2$-norm promotes the differentiability of the loss function and the smallest distance. As $\gamma$ grows, the function increases the sparsity of the counterfactuals and becomes harder to optimize. In the case of $\gamma=0$, the loss function converges to the $L_2$-norm, promoting the generation of counterfactuals with the smallest proximity.
    
    \begin{figure}[!h]
    \resizebox{\columnwidth}{!}{
    \includegraphics{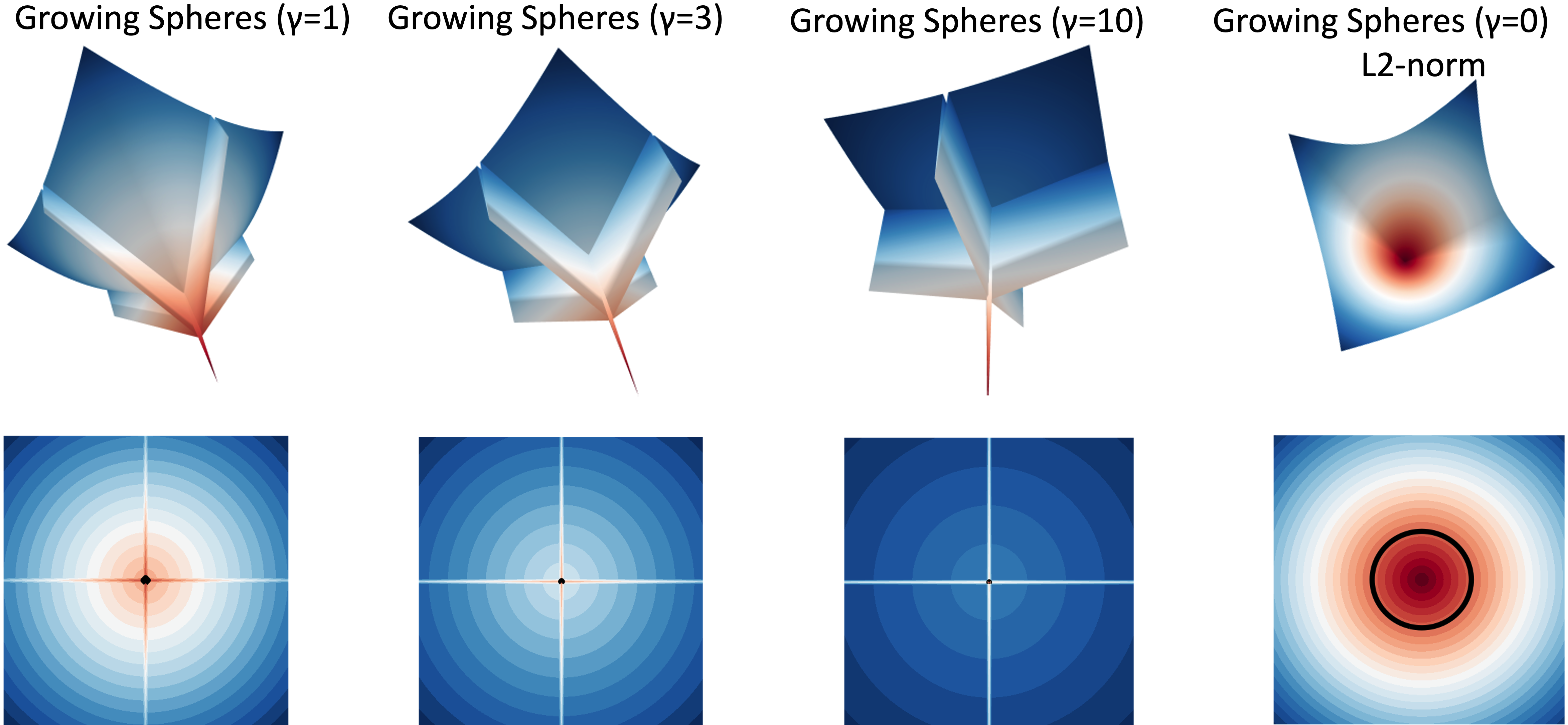}
    }
    \caption{Impact of the hyperparameter $\gamma$ in the Growing Spheres Loss function. When $\gamma=0$ the function collapses to the L$_2$-norm}
    \label{fig:GS_analysis}
    \end{figure}

    \vspace{0.2cm}
    \item \textbf{DiCE generates the best overall counterfactual explanations (both on numerical and mixed data)}.
    
    DiCE ensures \textit{plausibility}, and therefore, it can generate actionable and meaningful counterfactuals by constraining the space of sensitive features such as gender, race, etc. DiCE achieves this by a loss function that consists of a linear combination of three components: (1) a loss function to find a counterfactual candidate with a class different from the prediction of data instance $x$; (2) a proximity factor, which consists of normalizing the $L_1$-norm with the feature's median absolute deviation (initially proposed in WatcherCF~\cite{Watcher2018}) and (3) a diversity factor $dpp\_diversity$, which is computed using determinantal point processes~\cite{Mothilal2020DiCE}. 
    \begin{equation}
    x^* = \underset{x'\in X}{\operatorname{arg min}} \frac{1}{k}\sum_{i=1}^{k} yloss(f(x'),y) + \frac{\lambda_1}{k}\sum_{i=1}^{k} dist(x',x) - \lambda_2 \ ddp\_ diversity(x')
    \label{eq:dice}
    \end{equation}
    \begin{equation}
        dist(x', x) = \sum_{j=1} \frac{|c_i - x_j|}{MAD_j}, \text{~where~~}
    \end{equation}
    \[ MAD_j = median_{ i\in \{1,\dots,n\}}  \left|x_{i,j} - median_{l\in \{1,\dots,n\}}(x_l,j)\right|\]
    
    \vspace{0.2cm}
    We believe that DiCE did not achieve the best results regarding proximity metrics (compared to the GrowingSpheresCF algorithm), majorly because of its diversity component. The proximity measures become penalized by optimizing the loss function to generate diverse counterfactuals. Another aspect is due to immutable variables. Since DiCE ensures plausibility property, the generated counterfactuals may need larger values in other features to compensate for the constraints in certain variables. This will become clearer in this study when we present our counterfactual inspection and counterfactual inspection analysis (Section~\ref{sec:bias}). DiCE does not have a specific optimization term for sparsity in its loss function. Extending DiCE's loss function to incorporate the L$_0$-norm could improve 
    its performance in terms of proximity metrics.
    
    \vspace{0.2cm}
    \item \textbf{Decision Trees promote the best counterfactual explanations}.

    Decision trees learn by progressively splitting the feature space along several features to optimize the information gain, in other words, by minimizing the entropy. Additionally, contrary to neural networks, decision trees are deterministic, simplifying the feature selection process compared to other more complex models. Decision trees can easily generate counterfactuals by selecting an alternative splitting node from the feature space~\cite{Guidotti2019survey}. For instance, in Figure~\ref{fig:dt_analysis}, for a data instance $x$ with prediction $Y=0$, a counterfactual explanation may consist in the path $X_0=True \rightarrow X_1=True \rightarrow X_2 = True \rightarrow Y = 0$ (which is the counterfactual with the smallest sparsity, since only one node is changed), or the path $X_0 = True \rightarrow X_1 = False \rightarrow X_2 = False \rightarrow Y = 0$, or it can even generate the counterfactual $X_0 = False \rightarrow Y = 0$. Counterfactual explanation algorithms use different loss functions to compute a counterfactual explanation, which leads to different splitting sections of the tree.
    
    \begin{figure}[!h]
        \resizebox{\columnwidth}{!}{
        \includegraphics{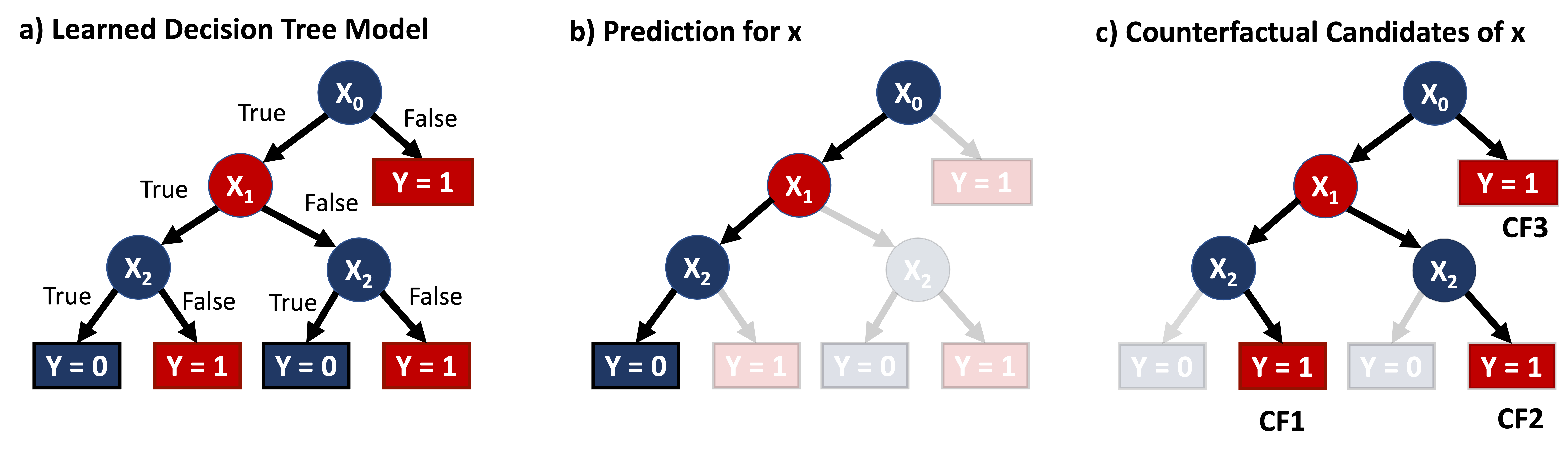}
        }
        \caption{Generation of counterfactual in Decision Trees. Panel a) shows a learned decision tree. Panel b) presents a prediction of a data instance $x$ for the same decision tree. Panel c) shows how counterfactual candidates can be generated from a decision tree easily and straightforwardly.}
        \label{fig:dt_analysis}
    \end{figure}
    
    
    
    \vspace{0.2cm}
    \item \textbf{WatcherCF generates the worst counterfactuals.}
    
    Watcher's algorithm generates counterfactual explanations by minimizing the distance between a data instance $x$ and a counterfactual candidate $x'$. This is achieved using the $L_1$-norm normalized by the inverse of the median absolute deviation of feature $j$. Since this distance function uses the $L_1$-norm, it induces sparsity in the counterfactual generation process (Equation~\ref{eq:watcher_distance1}).
        \begin{equation}
    d(x, x') = \sum^{p}_{j=1} \frac{|x_j - x'_j|}{MAD_j}, \text{~where~~}
            \label{eq:watcher_distance1} 
            \end{equation}
    \[ MAD_j = median_{ i\in	\{1,\dots,n\}}  \left|x_{i,j} - median_{l\in \{1,\dots,n\}}(x_l,j)\right|\]
    However, relying uniquely on the distance function is insufficient to ensure minimum proximity~\cite{Keane2021} since it can violate the feature space of sensitive features and lead to meaningless and biased counterfactuals. Additionally, as shown in the decision tree in Figure~\ref{fig:dt_analysis}, different counterfactuals can be generated with different interpretations of \textit{distances}. For instance, CF1 corresponds to the counterfactual with fewer feature changes (minimum sparsity) since it finds the opposite class in the same node as in the prediction of vector $x$. Another interpretation of distance is like in WatcherCF, where the algorithm uniquely focuses on finding the closest counterfactual with a different prediction of $x$. In a decision tree, this means that it would search for the first node split, leading to a different prediction. In Figure~\ref{fig:dt_analysis}, this corresponds to counterfactual CF3. Although this counterfactual corresponds to the closest path in the model that leads to an opposite class, it is also the counterfactual that would result in major feature changes and, therefore, an increased distance to the original input $x$. This analysis indicates that using a distance function uniquely to find counterfactuals does not guarantee plausibility and leads to biased and erroneous results. Therefore, such algorithms should not be used to generate counterfactual explanations, especially in scenarios of high-stakes decision-making~\cite{Rudin2019}.

    \begin{figure}[!h]
    \resizebox{\columnwidth}{!}{
    \includegraphics{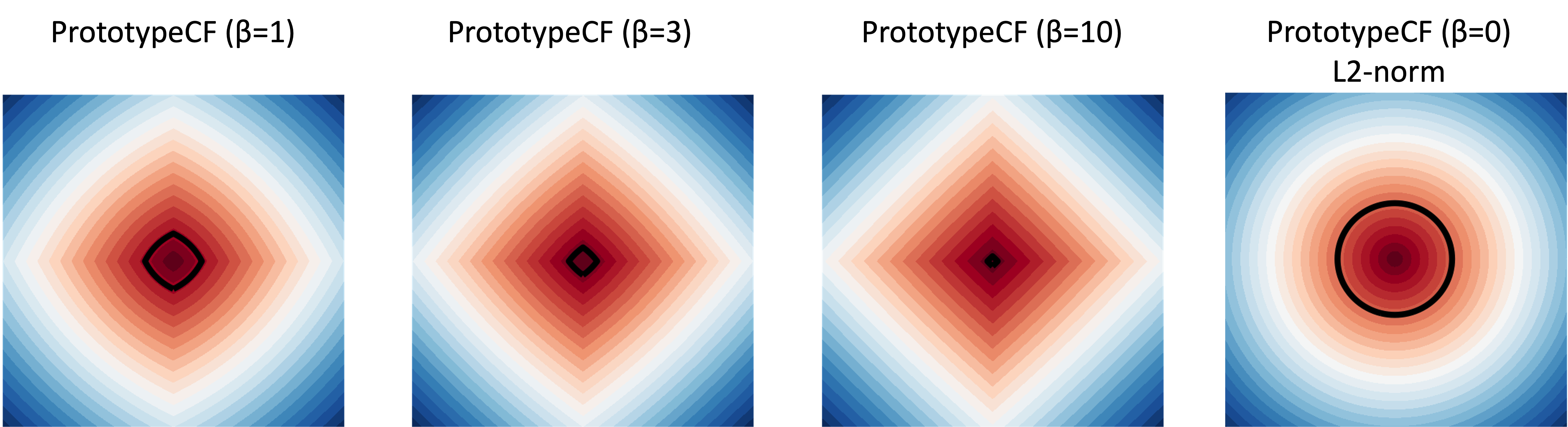}
    }
    \caption{Impact of the hyperparameter $\beta$ in the PrototypeCF Loss function. When $\beta=0$ the function collapses to the L$_2$-norm}
    \label{fig:Proto_analysis}
    \end{figure}
    
    \vspace{0.2cm}
    \item \textbf{Prototype generates the least efficient counterfactuals.}

    Prototype generates counterfactuals by determining the sufficient and minimum set of features required to predict the same class as the original instance and which features should be absent. The prototype's loss function consists of a weighted combination of five loss functions:
    \begin{equation}
            Loss(x,x') = c.L_{pred}(x,x') + Ldist(x,x') + L_{AE}(x,x')+L_{proto}(x,x'), \text{~~~where}
            \label{eq:loss_proto}
    \end{equation}
    \[ dist(x,x') = \beta \| x - x' \|_1 + \| x - x' \|_2  \]
    $L_{pred}$ measures the divergence between the class prediction probabilities, $L_1$ and $L_2$ correspond to the elastic net regularizer, $L_{AE}$ represents an autoencoder loss term that penalizes out-of-distribution counterfactual candidate instances (which can lead to uninterpretable counterfactuals). Finally, $L_{proto}$ guides the counterfactual search process toward a solution~\cite {Looveren2019}.

    We believe that the Prototype achieved worse efficiency because of the loss function $L_{AE}$, which trains an autoencoder each time a counterfactual is generated. This autoencoder performs worst in tree-like models, such as random forests.
    
    We also analyzed the impact of the hyperparameter $\beta$ in the computation of the distance function. Figure~\ref{fig:Proto_analysis} presents the evolution of the distance function for different values of $\beta$. When $\beta=1$, one can see that the function promotes sparsity and the smoothness of the L$_2$-norm that promotes differentiability (which is essential for the auto-encoder loss function). As $\beta$ grows, the function becomes 
    sparser and less differentiable. For the special case of $\beta=0$, the function collapses to the $L_{2}$-norm.
    
\end{itemize}

\subsection{Analysis of the Impact of Machine Learning Models in the Generation of Counterfactuals}

This section explores whether the underlying predictive model influences the quality of counterfactual generation.

\begin{figure}[!h]
    \resizebox{\columnwidth}{!}{
    \includegraphics{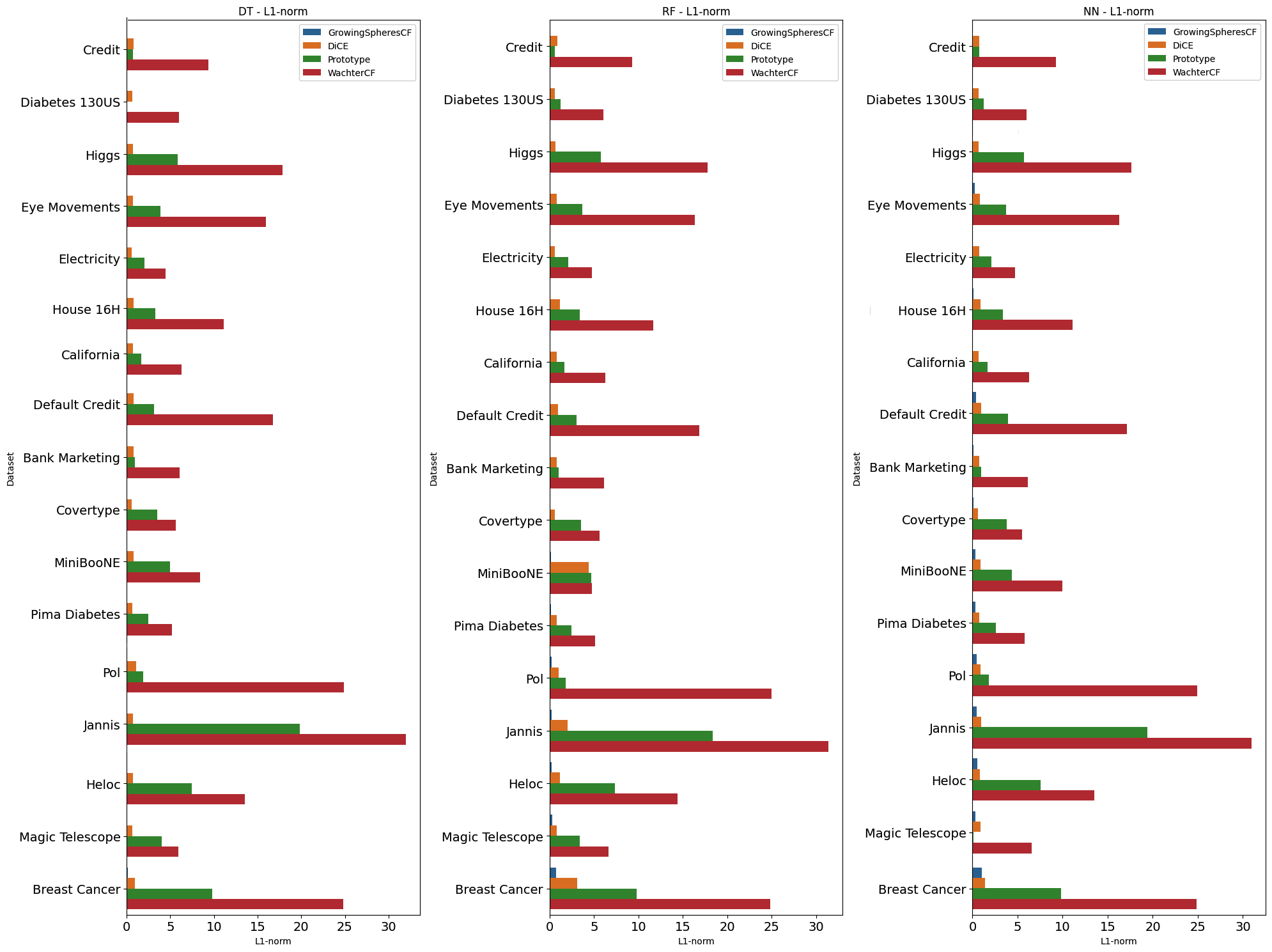}
    }
    \caption{L$_1$ Norm of the generated counterfactual explanations across the different machine learning algorithms: DT corresponds to a Decision Tree, RF to a Random Forest, and NN to a Neural Network.}
    \label{fig:proximity_L1}
\end{figure}

Figures~\ref{fig:proximity_L1} and \ref{fig:proximity_box_plot} present a comparison of the performance of each XAI counterfactual algorithm in terms of proximity for the different machine learning models. We found no significant impact of the nature of the predictive model (either a white box, a grey box, or a black box) on the quality of the counterfactual explanations. This is because the \textit{counterfactual generation process is always faithful to its underlying predictive model}.

\begin{figure}[!h]
    \resizebox{\columnwidth}{!}{
    \includegraphics{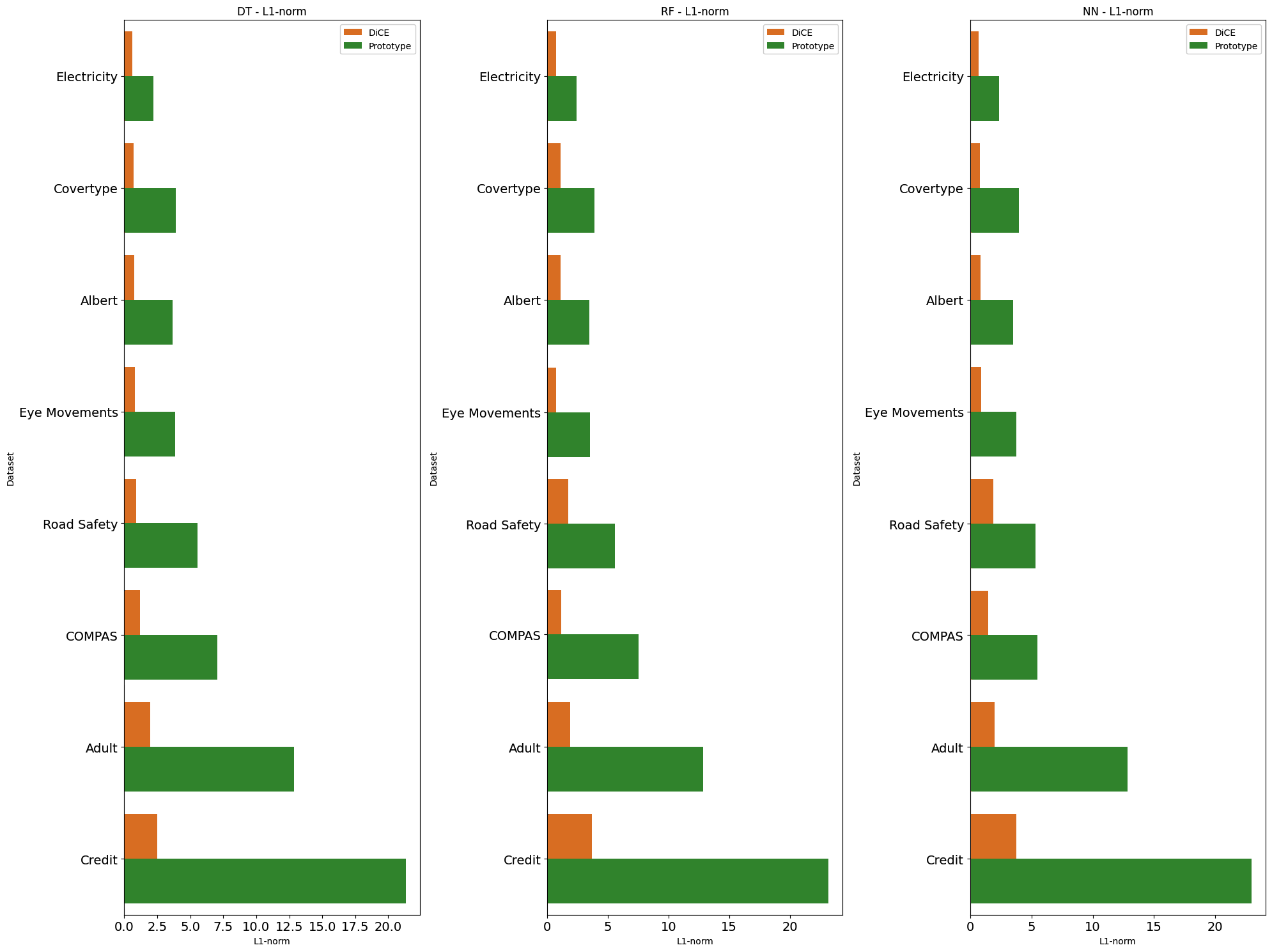}
    }
    \caption{L$_1$ Norm of the generated counterfactual explanations across the different machine learning algorithms: DT corresponds to a Decision Tree, RF to a Random Forest, and NN to a Neural Network.}
    \label{fig:proximity_L1_mixed}
\end{figure}

These results confirm that Decision Trees provide the best results across all counterfactual algorithms regarding L$_1$ norm (and L$_2$ norm). Additionally, one can see the consistency of WachterCF and Prototype since their median values are always mostly the same across the different ML algorithms. This is expected since these algorithms mostly find the same counterfactuals and do not offer diversity in the counterfactual generation process.
DiCE and GrowingSpheresCF, on the other hand, already show variability on the returned counterfactuals: DiCE has a diversity property that enables this, while GrowingSpheres is a more stochastic algorithm. Ultimately, one can see that the best-performing counterfactuals remain the best independently of the underlying ML model. The same applies to the worse-performing algorithms, which perform worse independently of the ML algorithm. This pattern underscores a key insight: when data is adequately represented and structured, leading to comparable performance across ML models, the specific choice of ML model becomes less critical. This is because these models reveal similar patterns within the data, placing the onus on the counterfactual algorithm to identify alterations in the input that would change the prediction outcome. While the decision boundaries of an ML algorithm can influence how a query is modified, the ultimate goal of these algorithms is to minimize distances, rendering the choice of ML model less consequential for the performance metrics typically used in XAI counterfactual analysis.

Conversely, counterfactual explanations operate differently than feature-attribution methods (such as LIME or SHAP). In general, post-hoc feature-attribution methods generate perturbations around the neighborhood of a local data instance that one wishes to explain. The perturbations and their predictions are fitted to a white-box model (such as linear regression). Because the white-box model is transparent, one can extract the weights of the features. Due to sampling and selection bias introduced by perturbation noise, the explanation's feature importance may not reflect the predictive model's feature importance. Therefore, the explanations may not be faithful~\cite{Bordt2022}. Counterfactuals are not generated based on permutation methods and are not focused on finding feature weights. They consist of optimization functions that follow the model's decision paths that lead to the desired outcomes. Therefore, \textit{the counterfactual explanation is always faithful}, and the underlying predictive model has little impact on the counterfactual generation process as demonstrated in Figure~\ref{fig:proximity_box_plot}.

\subsection{Analysis of the Impact of the Feature's Encoding Method}

\begin{figure}[!h]
    \includegraphics[scale=0.45]{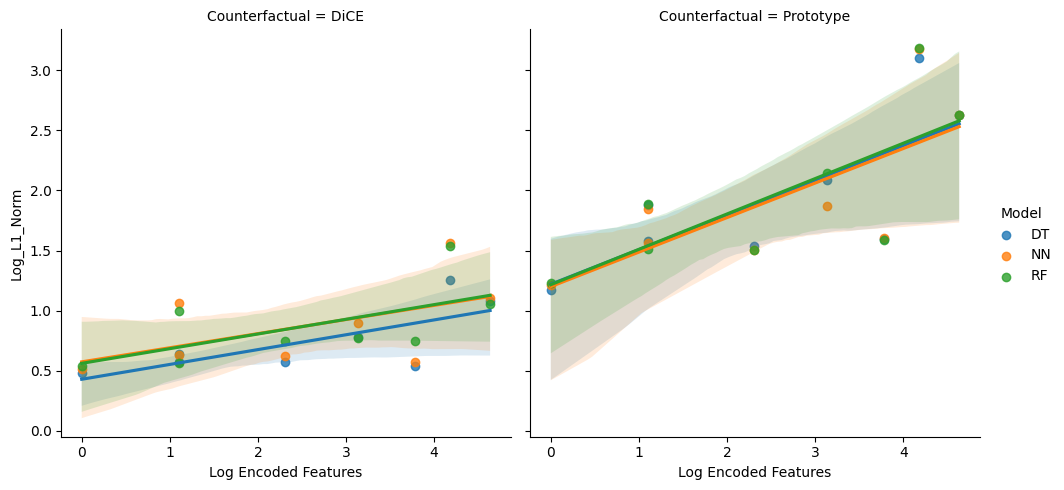}
    \caption{Regression analysis between the logarithm of the number of encoded features and the logarithm of the L$_1$-norm for the mixed datasets using DiCE and Prototype. In the figure, DT corresponds to the Decision Tree, RF to the Random Forest, and NN to the Neural Network. The plot shows that there is a positive correlation between the number of encoded features and L$_1$-norm.}
    \label{fig:reg_l1_norm}
\end{figure}

\begin{figure}[!h]
    \includegraphics[scale=0.45]{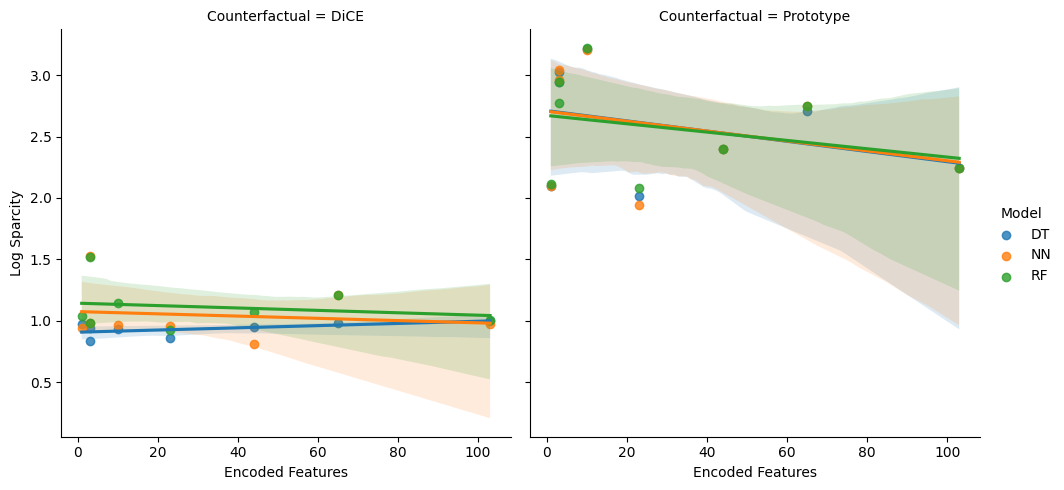}
    \caption{Regression analysis between the number of encoded features and the logarithm of the Sparsity for the mixed datasets using DiCE and Prototype. In the figure, DT corresponds to the Decision Tree, RF to the Random Forest, and NN to the Neural Network. The plot shows a significant correlation between sparsity and the number of features. }
    \label{fig:reg_spa}
\end{figure}

In our experiments, we have three datasets encoded with one hot encoding, which results in very sparse vectors. Additionally, we utilized the benchmark datasets from~\citet{grinsztajn2022tree}, which combine binary and ordinal encoding, translating into less sparse vectors. Figures~\ref{fig:reg_l1_norm} and \ref{fig:reg_spa} illustrate the impact of these different encoding methods on the performance of the algorithms under consideration in terms of L$_1$-norm and sparsity.

Figure~\ref{fig:reg_l1_norm} shows that for the DiCE algorithm, the positive coefficient indicates a direct relationship between the number of encoded features and the L1 norm. However, the relatively small magnitude of the regression coefficient (0.0153) suggests that this relationship is quite weak and statistically not significant. DiCE seems more robust to an increase in dataset complexity, maintaining a lower and more stable increase in the L{$_1$}-norm. On the other hand, Prototype presents a positive relationship between the number of encoded features and the L1 norm, however, with no statistical significance.

Figure \ref{fig:reg_spa} performs the same correlation analysis but with the Sparsity metric. DiCE shows a small negative relationship, suggesting that increases in the number of encoded features slightly decrease the sparsity of the counterfactuals generated by the DiCE algorithm. On the other hand, Prototype shows a more pronounced inverse relationship between the number of encoded features and sparsity. This suggests that the sparsity of counterfactual explanations decreases more significantly with increased encoded features.

To summarize, the choice of encoding method does not significantly impact the performance of counterfactual generation algorithms.

\section{Blind Reliance on Quantitative Metrics} \label{sec:bias}

In the previous section, DiCE and GrowingSpheresCF achieved the best results in proximity, sparsity, and functionality. However, GrowingSpheresCF does not ensure plausibility in the counterfactual generation process, while DiCE does. What are the consequences of not ensuring \textit{plausibility} in a counterfactual generation process? From the results alone of Tables~\ref{tab:num_data} and~\ref{tab:mixed_data}, one cannot see any difference, and the question cannot be answered because the quantitative approach alone does not provide any insights into whether an algorithm generates biased or unrealistic counterfactuals. This gap poses serious concerns in the XAI field because counterfactual algorithms that do not ensure plausibility may be more susceptible to generating inadequate explanations that may induce the user into error and cannot be detected using only quantitative metrics such as L$_p$ norms or sparsity.

\subsection{Experimental Setup}
In this section, we made a counterfactual inspection analysis to understand how different counterfactuals were generated using a decision tree as the predictive model. We chose the decision tree because it is a white box, and we can observe and inspect the different prediction paths. For each data instance $x$ of the test set, we used it as input to the decision tree model to predict its respective class, $y = f(x)$, and to generate its decision path, $Dt(x)$. We also generated the respective counterfactual explanation, $x'$, using an XAI counterfactual algorithm (such as DiCE, GrowingSpheresCF, Prototype, and Watcher) and also used it as input to the same decision tree to generate the prediction $y_c = f(x')$ and the decision path, $Dt(x')$. Finally, we compared both paths, focusing on decision nodes with sensible (immutable) variables (such as age or race). 

\subsection{Decision Path Analysis}

\begin{figure}[h!]
\resizebox{\columnwidth}{!}{
\includegraphics{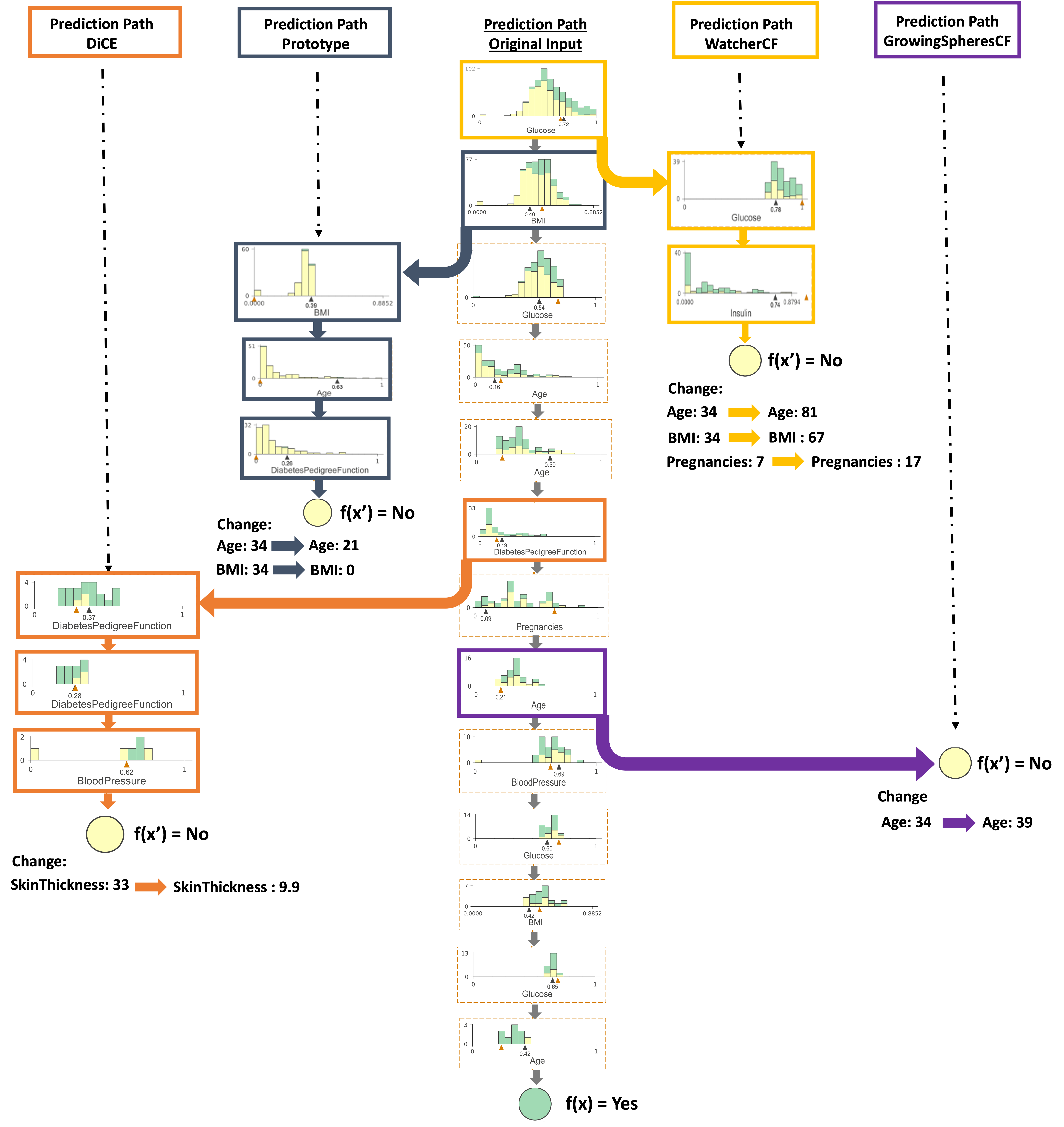}
}
\caption{Decision paths generated for a test data instance, $x$, and four counterfactual instances, $x'$, computed using DiCE, Prototype, Watcher, and GrowingSpheresCF for the Diabetes dataset. The figure shows the different notions of "minimum distance" between the algorithms. WatcherCF and Prototype tend to choose the first split of the decision path that leads to a prediction $f(x') = No$. GrowingSpheresCF and DiCE keep the minimum distance by following the decision path of $x$ until they find the closest branch split that leads to $f(x')=No$. Since DiCE ensures plausibility, it must choose other branches that do not contain immutable variables, leading to counterfactuals that do not necessarily have the smallest distance.}
\label{fig:bias}
\end{figure}

A good counterfactual should produce the closest counterfactual result with the smallest change, which means the decision paths, $Dt(x)$ and $Dt(x')$, should be comparable. In Figure~\ref{fig:bias}, we observe that GrowingSpheresCF follows the same route as the decision path of $Dt(x)$. However, it starts to diverge at depth $7$, where the algorithm finds a branch in the decision tree that enables it to reach the end of the path in the desired counterfactual prediction (Diabetes = No). To do this split, GrowingSpheresCF changed the variable $Age: 34 \rightarrow 39$, suggesting that for a person to \textit{not have diabetes, the person needs to get $5$ years older}. Although GrowingSpheresCF achieved the best proximity and sparsity performance (Table~\ref{tab:num_data}), this counterfactual explanation is erroneous and is a consequence of the algorithm's inability to satisfy the plausibility property.

DiCE, on the other hand, can constrain features in its optimization process. By incorporating domain knowledge about immutable variables (in this case, pregnancies and age), one can also note that DiCE's $Dt(x') \approx Dt(x)$ and splits at depth $5$ because it would lead to the smallest decision path containing the counterfactual prediction $f(x') = No$ using no immutable variables. Enforcing plausibility does not necessarily translate into counterfactuals with the smallest proximity metric. This reinforces our findings that proximity metrics alone should not be used in counterfactuals that do not contain plausibility mechanisms in their formalizations. However, it does promote explanations that are more faithful to the domain knowledge and are more human-understandable. DiCE's counterfactual for this example suggests that \textit{if the person had a SkinThickness change from} $SkinThickness: 33 \rightarrow 9.9$, \textit{she would not have diabetes}. According to domain knowledge, skin thickness correlates with insulin resistance, which is a potential indicator of diabetes.

WatcherCF and Prototype present similar search patterns. In Figure~\ref{fig:bias}, one can notice that WatcherCF tends to optimize the counterfactual search by looking at the closest point to $x$ that would promote a counterfactual outcome. This means it searches for counterfactuals close to the decision boundary. For the example in Figure~\ref{fig:bias}, WatcherCF finds the first split node from $Dt(x)$ that can lead to the desired outcome, irrespective of the feature range change. According to the underlying decision tree model, this path is indeed the closest one that leads to the counterfactual outcome, but it is not necessarily the one that leads to the smallest $d(x,x')$. This analysis can complement the results obtained in Table~\ref{tab:num_data}, where WatcherCF obtained the worst results in terms of proximity. However, its loss function is focused on optimizing proximity, $argmin_{x'}~d(x,x')$. Although it seems counterintuitive, one can understand from this analysis that the optimization of minimum distance is grounded on the predictive model. Therefore, the minimization will follow decision paths that lead to desired outcomes, even if that translates into a significant feature change. In Figure~\ref{fig:bias}, the counterfactual found by WatcherCF suggests changing $Age: 34 \rightarrow Age: 81$, $BMI: 34 \rightarrow 67$, and $Pregnancies: 7 \rightarrow Pregnancies: 17$, translating into an explanation that states that \textit{if a person gets 47 years older, increases BMI to 67, and has an additional 10 children, then she would not have diabetes}, which is a nonsensical statement according to the domain knowledge. 

For the curious reader, we provide the functions to generate these decision paths for different datasets, ML models, and counterfactuals in our public benchmark repository. One can easily generate more examples like the one presented in Figure~\ref{fig:bias}.

\subsection{Summary and Final Discussion}

This study focuses on model-agnostic, instance-centric, explainable algorithms. We have specifically selected these algorithms because they are the most widely used in the literature and provide reproducible results due to their open-source nature. This choice allows us to maintain a narrow and targeted scope for our research, ensuring that our findings are reliable and valid for this specific class of algorithms, enabling us to dive deeper into the implications of each algorithm. Indeed, one of the major contributions of this paper, when compared to other benchmarks in the depth of our analysis, led us to unique conclusions about the usefulness of functionally grounded evaluation metrics for counterfactual explanations.

This benchmark comprises stationary, well-represented, and independently and identically distributed (iid) data. Under these conditions, regardless of the underlying machine-learning algorithm, similar patterns in the data will always be learned. Consequently, including more machine learning models would not alter or contribute additional insight to our central argument: "functionally grounded evaluation metrics alone are not enough to express what constitutes a good counterfactual".

Our study found that properly imposing domain knowledge in the counterfactual generation process can penalize its proximity metric to the query datapoint, which is a good thing. Thus, including more counterfactual models and methods (risking falling outside of the scope of this work) would not change this result and might divert attention from this key finding. 

\section{Conclusion}

This study used state-of-the-art quantitative metrics to present a benchmark of several model-agnostic counterfactual algorithms. As part of this evaluation, we also examined decision-tree paths to understand the structure of the counterfactual explanations. Our main findings suggest that relying solely on quantitative metrics, such as proximity or sparsity, is insufficient and a poor indicator of a counterfactual explanation's quality. Counterfactuals that do not ensure plausibility and do not capture domain knowledge may have good proximity scores, but their explanations may induce users into falsely believing in a decision~\cite{Bordt2022}. This means that explainable counterfactual algorithms that do not consider plausibility in their internal mechanisms cannot be evaluated with the current state-of-the-art evaluation metrics, and their results may be biased.

This study also demonstrated the advantages of inspecting generated counterfactual explanations by analyzing their decision paths. We recommend future research to develop a more robust 
inspection of counterfactual explanations to complement the quantitative metrics. Generating counterfactuals' decision paths can provide novel insights into the counterfactual generation process, provide a plausibility analysis, or even verify alignment with context and domain knowledge. 

This work also enabled us to investigate different predictive models and their role in the counterfactual generation. We found that the predictive model (either a white box, a grey box, or a black box) did not significantly impact the generation of counterfactuals.

Regarding the specific properties of XAI counterfactual algorithms, we found that relying uniquely on quantitative metrics such as proximity and sparsity, GrowingSpheresCF achieved the best results because its loss function optimizes both sparsity (through L$_0$-norm) and distance (through $L_2$-norm). However, this algorithm does not satisfy the plausibility property, and consequently, it may generate biased and erroneous counterfactual explanations. We consider that DiCE achieved the best outcomes in both quantitative and qualitative aspects because it achieved very good proximity results and ensured plausibility, which means that the counterfactual generation process considers immutable variables. Finally, WatcherCF is the least-performing algorithm, generating highly biased counterfactuals by 
solely relying on a distance function. 
As presented in this study, WatcherCF
always chose the first splitting node of the decision tree 
even if this path deviated significantly from the original input. 

We end this study with the observation that the current scientific literature is facing a replication crisis~\cite{Lawless2019} and still needs a standardized evaluation framework together with guidelines and recommendations to correctly evaluate the quality of the counterfactual generation algorithm and the quality of the counterfactual explanations~\cite{Bordt2022}. As we have seen in this study, "good" counterfactual generation algorithms do not necessarily promote "good" counterfactual explanations. The scientific community needs an evaluation framework that can promote a fair comparison of XAI counterfactual algorithms to promote reproducible scientific research. Additionally, we must incorporate user studies into our evaluation framework for future work. These studies would involve presenting generated counterfactuals to end-users and soliciting their feedback on their plausibility and actionability. 
Undeniably, the profound importance of user-centric evaluations cannot be overstated. These assessments serve as an essential conduit, channeling invaluable insights regarding the tangible effectiveness of counterfactual descriptions within the realities of our world. They illuminate a path toward a more nuanced, refined understanding of what embodies exemplary counterfactual explanations.

\section*{Acknowledgements}
This work was partially supported by national funds through FCT, \textit{Fundação para a Ciência e a Tecnologia} project UIDB/50021/2020 (DOI:10.54499/UIDB/50021/2020) and 2022.09212.PTDC (DOI: 10.54499/2022.09212.PTDC) under the auspices of the UNESCO Chair on AI\&VR of the University of Lisbon.

\bibliographystyle{ACM-Reference-Format}

\appendix

\section{Performance of Machine Learning Algorithms}
Table~\ref{tab:classification_metrics} presents the accuracy, precision, recall, and F1-score results. We trained the ML algorithms so that they would perform similarly. The hyperparameters used in each model can be found in our public repository (\url{https://tinyurl.com/4tyakw98}).

\begin{table}[h!]
\caption{Overall performance achieved using different types of machine learning algorithms: a decision tree (white box), a random forest (grey box), and a neural network (black box) over different datasets (for details on the datasets, please check \citet{grinsztajn2022tree}). In the table, \textit{Acc} stands for model Accuracy, \textit{Prec} for model Precision, \textit{Rec} for model Recall, and \textit{F1} for F1 score. }
\label{tab:classification_metrics}
\resizebox{\columnwidth}{!}{
\begin{tabular}{lllllllllllll}
\hline
\multicolumn{1}{|l|}{} &
  \multicolumn{4}{c|}{\textbf{Decision Tree}} &
  \multicolumn{4}{c|}{\textbf{Random Forest}} &
  \multicolumn{4}{c|}{\textbf{Neural Net}} \\ \cline{2-13} 
\multicolumn{1}{|l|}{\multirow{-2}{*}{\textbf{Dataset}}} &
  \multicolumn{1}{l|}{\textbf{Acc}} &
  \multicolumn{1}{l|}{\textbf{Prec}} &
  \multicolumn{1}{l|}{\textbf{Rec}} &
  \multicolumn{1}{l|}{\textbf{F1}} &
  \multicolumn{1}{l|}{\textbf{Acc}} &
  \multicolumn{1}{l|}{\textbf{Prec}} &
  \multicolumn{1}{l|}{\textbf{Rec}} &
  \multicolumn{1}{l|}{\textbf{F1}} &
  \multicolumn{1}{l|}{\textbf{Acc}} &
  \multicolumn{1}{l|}{\textbf{Prec}} &
  \multicolumn{1}{l|}{\textbf{Rec}} &
  \multicolumn{1}{l|}{\textbf{F1}} \\ \hline
\rowcolor[HTML]{EFEFEF} 
\multicolumn{1}{|l|}{\cellcolor[HTML]{EFEFEF}Electricity (mixed)} &
  \multicolumn{1}{l|}{\cellcolor[HTML]{EFEFEF}0.8456} &
  \multicolumn{1}{l|}{\cellcolor[HTML]{EFEFEF}0.8483} &
  \multicolumn{1}{l|}{\cellcolor[HTML]{EFEFEF}0.8458} &
  \multicolumn{1}{l|}{\cellcolor[HTML]{EFEFEF}0.8470} &
  \multicolumn{1}{l|}{\cellcolor[HTML]{EFEFEF}0.8835} &
  \multicolumn{1}{l|}{\cellcolor[HTML]{EFEFEF}0.8861} &
  \multicolumn{1}{l|}{\cellcolor[HTML]{EFEFEF}0.8830} &
  \multicolumn{1}{l|}{\cellcolor[HTML]{EFEFEF}0.8845} &
  \multicolumn{1}{l|}{\cellcolor[HTML]{EFEFEF}0.7659} &
  \multicolumn{1}{l|}{\cellcolor[HTML]{EFEFEF}0.7742} &
  \multicolumn{1}{l|}{\cellcolor[HTML]{EFEFEF}0.7579} &
  \multicolumn{1}{l|}{\cellcolor[HTML]{EFEFEF}0.7660} \\ \hline
\multicolumn{1}{|l|}{Eye Movements (mixed)} &
  \multicolumn{1}{l|}{0.5575} &
  \multicolumn{1}{l|}{0.5705} &
  \multicolumn{1}{l|}{0.5386} &
  \multicolumn{1}{l|}{0.5541} &
  \multicolumn{1}{l|}{0.6101} &
  \multicolumn{1}{l|}{0.6250} &
  \multicolumn{1}{l|}{0.5901} &
  \multicolumn{1}{l|}{0.6071} &
  \multicolumn{1}{l|}{0.5783} &
  \multicolumn{1}{l|}{0.6025} &
  \multicolumn{1}{l|}{0.5107} &
  \multicolumn{1}{l|}{0.5528} \\ \hline
\rowcolor[HTML]{EFEFEF} 
\multicolumn{1}{|l|}{\cellcolor[HTML]{EFEFEF}Covertype mixed} &
  \multicolumn{1}{l|}{\cellcolor[HTML]{EFEFEF}0.9297} &
  \multicolumn{1}{l|}{\cellcolor[HTML]{EFEFEF}0.9283} &
  \multicolumn{1}{l|}{\cellcolor[HTML]{EFEFEF}0.9311} &
  \multicolumn{1}{l|}{\cellcolor[HTML]{EFEFEF}0.9297} &
  \multicolumn{1}{l|}{\cellcolor[HTML]{EFEFEF}0.9507} &
  \multicolumn{1}{l|}{\cellcolor[HTML]{EFEFEF}0.9568} &
  \multicolumn{1}{l|}{\cellcolor[HTML]{EFEFEF}0.9439} &
  \multicolumn{1}{l|}{\cellcolor[HTML]{EFEFEF}0.9503} &
  \multicolumn{1}{l|}{\cellcolor[HTML]{EFEFEF}0.8316} &
  \multicolumn{1}{l|}{\cellcolor[HTML]{EFEFEF}0.8027} &
  \multicolumn{1}{l|}{\cellcolor[HTML]{EFEFEF}0.8785} &
  \multicolumn{1}{l|}{\cellcolor[HTML]{EFEFEF}0.8389} \\ \hline
\multicolumn{1}{|l|}{Albert (mixed)} &
  \multicolumn{1}{l|}{0.5633} &
  \multicolumn{1}{l|}{0.5607} &
  \multicolumn{1}{l|}{0.5715} &
  \multicolumn{1}{l|}{0.5660} &
  \multicolumn{1}{l|}{0.6500} &
  \multicolumn{1}{l|}{0.6442} &
  \multicolumn{1}{l|}{0.6648} &
  \multicolumn{1}{l|}{0.6543} &
  \multicolumn{1}{l|}{0.6394} &
  \multicolumn{1}{l|}{0.6335} &
  \multicolumn{1}{l|}{0.6559} &
  \multicolumn{1}{l|}{0.6445} \\ \hline
\rowcolor[HTML]{EFEFEF} 
\multicolumn{1}{|l|}{\cellcolor[HTML]{EFEFEF}Road safety (mixed)} &
  \multicolumn{1}{l|}{\cellcolor[HTML]{EFEFEF}0.7222} &
  \multicolumn{1}{l|}{\cellcolor[HTML]{EFEFEF}0.7271} &
  \multicolumn{1}{l|}{\cellcolor[HTML]{EFEFEF}0.7236} &
  \multicolumn{1}{l|}{\cellcolor[HTML]{EFEFEF}0.7253} &
  \multicolumn{1}{l|}{\cellcolor[HTML]{EFEFEF}0.7844} &
  \multicolumn{1}{l|}{\cellcolor[HTML]{EFEFEF}0.7771} &
  \multicolumn{1}{l|}{\cellcolor[HTML]{EFEFEF}0.8060} &
  \multicolumn{1}{l|}{\cellcolor[HTML]{EFEFEF}0.7913} &
  \multicolumn{1}{l|}{\cellcolor[HTML]{EFEFEF}0.7668} &
  \multicolumn{1}{l|}{\cellcolor[HTML]{EFEFEF}0.7717} &
  \multicolumn{1}{l|}{\cellcolor[HTML]{EFEFEF}0.7667} &
  \multicolumn{1}{l|}{\cellcolor[HTML]{EFEFEF}0.7692} \\ \hline
\multicolumn{1}{|l|}{California (num)} &
  \multicolumn{1}{l|}{0.8098} &
  \multicolumn{1}{l|}{0.8097} &
  \multicolumn{1}{l|}{0.8193} &
  \multicolumn{1}{l|}{0.8145} &
  \multicolumn{1}{l|}{0.8764} &
  \multicolumn{1}{l|}{0.8912} &
  \multicolumn{1}{l|}{0.8629} &
  \multicolumn{1}{l|}{0.8768} &
  \multicolumn{1}{l|}{0.8251} &
  \multicolumn{1}{l|}{0.8758} &
  \multicolumn{1}{l|}{0.7655} &
  \multicolumn{1}{l|}{0.8169} \\ \hline
\rowcolor[HTML]{EFEFEF} 
\multicolumn{1}{|l|}{\cellcolor[HTML]{EFEFEF}Credit (num)} &
  \multicolumn{1}{l|}{\cellcolor[HTML]{EFEFEF}0.6909} &
  \multicolumn{1}{l|}{\cellcolor[HTML]{EFEFEF}0.6881} &
  \multicolumn{1}{l|}{\cellcolor[HTML]{EFEFEF}0.6963} &
  \multicolumn{1}{l|}{\cellcolor[HTML]{EFEFEF}0.6922} &
  \multicolumn{1}{l|}{\cellcolor[HTML]{EFEFEF}0.7822} &
  \multicolumn{1}{l|}{\cellcolor[HTML]{EFEFEF}0.7889} &
  \multicolumn{1}{l|}{\cellcolor[HTML]{EFEFEF}0.7692} &
  \multicolumn{1}{l|}{\cellcolor[HTML]{EFEFEF}0.7790} &
  \multicolumn{1}{l|}{\cellcolor[HTML]{EFEFEF}0.7478} &
  \multicolumn{1}{l|}{\cellcolor[HTML]{EFEFEF}0.8377} &
  \multicolumn{1}{l|}{\cellcolor[HTML]{EFEFEF}0.6134} &
  \multicolumn{1}{l|}{\cellcolor[HTML]{EFEFEF}0.7082} \\ \hline
\multicolumn{1}{|l|}{Heloc (num)} &
  \multicolumn{1}{l|}{0.6325} &
  \multicolumn{1}{l|}{0.6613} &
  \multicolumn{1}{l|}{0.5987} &
  \multicolumn{1}{l|}{0.6285} &
  \multicolumn{1}{l|}{0.6992} &
  \multicolumn{1}{l|}{0.7404} &
  \multicolumn{1}{l|}{0.6501} &
  \multicolumn{1}{l|}{0.6923} &
  \multicolumn{1}{l|}{0.6992} &
  \multicolumn{1}{l|}{0.7408} &
  \multicolumn{1}{l|}{0.6469} &
  \multicolumn{1}{l|}{0.6907} \\ \hline
\rowcolor[HTML]{EFEFEF} 
\multicolumn{1}{|l|}{\cellcolor[HTML]{EFEFEF}Jannis (num)} &
  \multicolumn{1}{l|}{\cellcolor[HTML]{EFEFEF}0.6894} &
  \multicolumn{1}{l|}{\cellcolor[HTML]{EFEFEF}0.6899} &
  \multicolumn{1}{l|}{\cellcolor[HTML]{EFEFEF}0.6855} &
  \multicolumn{1}{l|}{\cellcolor[HTML]{EFEFEF}0.6877} &
  \multicolumn{1}{l|}{\cellcolor[HTML]{EFEFEF}0.7562} &
  \multicolumn{1}{l|}{\cellcolor[HTML]{EFEFEF}0.7608} &
  \multicolumn{1}{l|}{\cellcolor[HTML]{EFEFEF}0.8222} &
  \multicolumn{1}{l|}{\cellcolor[HTML]{EFEFEF}0.7903} &
  \multicolumn{1}{l|}{\cellcolor[HTML]{EFEFEF}0.7562} &
  \multicolumn{1}{l|}{\cellcolor[HTML]{EFEFEF}0.7373} &
  \multicolumn{1}{l|}{\cellcolor[HTML]{EFEFEF}0.7940} &
  \multicolumn{1}{l|}{\cellcolor[HTML]{EFEFEF}0.7646} \\ \hline
\multicolumn{1}{|l|}{Diabetes130US (num)} &
  \multicolumn{1}{l|}{0.5342} &
  \multicolumn{1}{l|}{0.5354} &
  \multicolumn{1}{l|}{0.5080} &
  \multicolumn{1}{l|}{0.5213} &
  \multicolumn{1}{l|}{0.6421} &
  \multicolumn{1}{l|}{0.5510} &
  \multicolumn{1}{l|}{0.5552} &
  \multicolumn{1}{l|}{0.5531} &
  \multicolumn{1}{l|}{0.6049} &
  \multicolumn{1}{l|}{0.6421} &
  \multicolumn{1}{l|}{0.4716} &
  \multicolumn{1}{l|}{0.5438} \\ \hline
\rowcolor[HTML]{EFEFEF} 
\multicolumn{1}{|l|}{\cellcolor[HTML]{EFEFEF}Eye Movements (num)} &
  \multicolumn{1}{l|}{\cellcolor[HTML]{EFEFEF}0.5564} &
  \multicolumn{1}{l|}{\cellcolor[HTML]{EFEFEF}0.5645} &
  \multicolumn{1}{l|}{\cellcolor[HTML]{EFEFEF}0.5730} &
  \multicolumn{1}{l|}{\cellcolor[HTML]{EFEFEF}0.5687} &
  \multicolumn{1}{l|}{\cellcolor[HTML]{EFEFEF}0.6142} &
  \multicolumn{1}{l|}{\cellcolor[HTML]{EFEFEF}0.6211} &
  \multicolumn{1}{l|}{\cellcolor[HTML]{EFEFEF}0.5944} &
  \multicolumn{1}{l|}{\cellcolor[HTML]{EFEFEF}0.6075} &
  \multicolumn{1}{l|}{\cellcolor[HTML]{EFEFEF}0.5564} &
  \multicolumn{1}{l|}{\cellcolor[HTML]{EFEFEF}0.6142} &
  \multicolumn{1}{l|}{\cellcolor[HTML]{EFEFEF}0.3519} &
  \multicolumn{1}{l|}{\cellcolor[HTML]{EFEFEF}0.4475} \\ \hline
\multicolumn{1}{|l|}{Higgs (num)} &
  \multicolumn{1}{l|}{0.6406} &
  \multicolumn{1}{l|}{0.6420} &
  \multicolumn{1}{l|}{0.6385} &
  \multicolumn{1}{l|}{0.6402} &
  \multicolumn{1}{l|}{0.7300} &
  \multicolumn{1}{l|}{0.7291} &
  \multicolumn{1}{l|}{0.7334} &
  \multicolumn{1}{l|}{0.7313} &
  \multicolumn{1}{l|}{0.7128} &
  \multicolumn{1}{l|}{0.7168} &
  \multicolumn{1}{l|}{0.7053} &
  \multicolumn{1}{l|}{0.7110} \\ \hline
\rowcolor[HTML]{EFEFEF} 
\multicolumn{1}{|l|}{\cellcolor[HTML]{EFEFEF}Default of Credit (num)} &
  \multicolumn{1}{l|}{\cellcolor[HTML]{EFEFEF}0.6146} &
  \multicolumn{1}{l|}{\cellcolor[HTML]{EFEFEF}0.6161} &
  \multicolumn{1}{l|}{\cellcolor[HTML]{EFEFEF}0.6269} &
  \multicolumn{1}{l|}{\cellcolor[HTML]{EFEFEF}0.6215} &
  \multicolumn{1}{l|}{\cellcolor[HTML]{EFEFEF}0.7163} &
  \multicolumn{1}{l|}{\cellcolor[HTML]{EFEFEF}0.7604} &
  \multicolumn{1}{l|}{\cellcolor[HTML]{EFEFEF}0.6393} &
  \multicolumn{1}{l|}{\cellcolor[HTML]{EFEFEF}0.6946} &
  \multicolumn{1}{l|}{\cellcolor[HTML]{EFEFEF}0.7012} &
  \multicolumn{1}{l|}{\cellcolor[HTML]{EFEFEF}0.8037} &
  \multicolumn{1}{l|}{\cellcolor[HTML]{EFEFEF}0.5398} &
  \multicolumn{1}{l|}{\cellcolor[HTML]{EFEFEF}0.6458} \\ \hline
\multicolumn{1}{|l|}{MiniBooNE (num)} &
  \multicolumn{1}{l|}{0.8748} &
  \multicolumn{1}{l|}{0.8753} &
  \multicolumn{1}{l|}{0.8723} &
  \multicolumn{1}{l|}{0.8738} &
  \multicolumn{1}{l|}{0.9258} &
  \multicolumn{1}{l|}{0.9101} &
  \multicolumn{1}{l|}{0.9440} &
  \multicolumn{1}{l|}{0.9267} &
  \multicolumn{1}{l|}{0.7855} &
  \multicolumn{1}{l|}{0.8682} &
  \multicolumn{1}{l|}{0.6703} &
  \multicolumn{1}{l|}{0.7565} \\ \hline
\rowcolor[HTML]{EFEFEF} 
\multicolumn{1}{|l|}{\cellcolor[HTML]{EFEFEF}Bank Marketing (num)} &
  \multicolumn{1}{l|}{\cellcolor[HTML]{EFEFEF}0.7449} &
  \multicolumn{1}{l|}{\cellcolor[HTML]{EFEFEF}0.7574} &
  \multicolumn{1}{l|}{\cellcolor[HTML]{EFEFEF}0.7412} &
  \multicolumn{1}{l|}{\cellcolor[HTML]{EFEFEF}0.7492} &
  \multicolumn{1}{l|}{\cellcolor[HTML]{EFEFEF}0.8165} &
  \multicolumn{1}{l|}{\cellcolor[HTML]{EFEFEF}0.8221} &
  \multicolumn{1}{l|}{\cellcolor[HTML]{EFEFEF}0.8208} &
  \multicolumn{1}{l|}{\cellcolor[HTML]{EFEFEF}0.8215} &
  \multicolumn{1}{l|}{\cellcolor[HTML]{EFEFEF}0.8024} &
  \multicolumn{1}{l|}{\cellcolor[HTML]{EFEFEF}0.7680} &
  \multicolumn{1}{l|}{\cellcolor[HTML]{EFEFEF}0.8821} &
  \multicolumn{1}{l|}{\cellcolor[HTML]{EFEFEF}0.8211} \\ \hline
\multicolumn{1}{|l|}{Magic Telescope (num)} &
  \multicolumn{1}{l|}{0.7713} &
  \multicolumn{1}{l|}{0.7655} &
  \multicolumn{1}{l|}{0.7836} &
  \multicolumn{1}{l|}{0.7744} &
  \multicolumn{1}{l|}{0.8467} &
  \multicolumn{1}{l|}{0.8633} &
  \multicolumn{1}{l|}{0.8246} &
  \multicolumn{1}{l|}{0.8435} &
  \multicolumn{1}{l|}{0.7950} &
  \multicolumn{1}{l|}{0.8874} &
  \multicolumn{1}{l|}{0.6766} &
  \multicolumn{1}{l|}{0.7678} \\ \hline
\rowcolor[HTML]{EFEFEF} 
\multicolumn{1}{|l|}{\cellcolor[HTML]{EFEFEF}House 16H (num)} &
  \multicolumn{1}{l|}{\cellcolor[HTML]{EFEFEF}0.7937} &
  \multicolumn{1}{l|}{\cellcolor[HTML]{EFEFEF}0.7793} &
  \multicolumn{1}{l|}{\cellcolor[HTML]{EFEFEF}0.8030} &
  \multicolumn{1}{l|}{\cellcolor[HTML]{EFEFEF}0.7910} &
  \multicolumn{1}{l|}{\cellcolor[HTML]{EFEFEF}0.8660} &
  \multicolumn{1}{l|}{\cellcolor[HTML]{EFEFEF}0.8545} &
  \multicolumn{1}{l|}{\cellcolor[HTML]{EFEFEF}0.8729} &
  \multicolumn{1}{l|}{\cellcolor[HTML]{EFEFEF}0.8636} &
  \multicolumn{1}{l|}{\cellcolor[HTML]{EFEFEF}0.8443} &
  \multicolumn{1}{l|}{\cellcolor[HTML]{EFEFEF}0.8064} &
  \multicolumn{1}{l|}{\cellcolor[HTML]{EFEFEF}0.8945} &
  \multicolumn{1}{l|}{\cellcolor[HTML]{EFEFEF}0.8482} \\ \hline
\multicolumn{1}{|l|}{Pol (num)} &
  \multicolumn{1}{l|}{0.9719} &
  \multicolumn{1}{l|}{0.9783} &
  \multicolumn{1}{l|}{0.9654} &
  \multicolumn{1}{l|}{0.9718} &
  \multicolumn{1}{l|}{0.9860} &
  \multicolumn{1}{l|}{0.9917} &
  \multicolumn{1}{l|}{0.9802} &
  \multicolumn{1}{l|}{0.9859} &
  \multicolumn{1}{l|}{0.9777} &
  \multicolumn{1}{l|}{0.9618} &
  \multicolumn{1}{l|}{0.9951} &
  \multicolumn{1}{l|}{0.9781} \\ \hline
\rowcolor[HTML]{EFEFEF} 
\multicolumn{1}{|l|}{\cellcolor[HTML]{EFEFEF}Covertype (num)} &
  \multicolumn{1}{l|}{\cellcolor[HTML]{EFEFEF}0.9092} &
  \multicolumn{1}{l|}{\cellcolor[HTML]{EFEFEF}0.9094} &
  \multicolumn{1}{l|}{\cellcolor[HTML]{EFEFEF}0.9105} &
  \multicolumn{1}{l|}{\cellcolor[HTML]{EFEFEF}0.9099} &
  \multicolumn{1}{l|}{\cellcolor[HTML]{EFEFEF}0.9427} &
  \multicolumn{1}{l|}{\cellcolor[HTML]{EFEFEF}0.9331} &
  \multicolumn{1}{l|}{\cellcolor[HTML]{EFEFEF}0.9549} &
  \multicolumn{1}{l|}{\cellcolor[HTML]{EFEFEF}0.9439} &
  \multicolumn{1}{l|}{\cellcolor[HTML]{EFEFEF}0.7931} &
  \multicolumn{1}{l|}{\cellcolor[HTML]{EFEFEF}0.7650} &
  \multicolumn{1}{l|}{\cellcolor[HTML]{EFEFEF}0.8508} &
  \multicolumn{1}{l|}{\cellcolor[HTML]{EFEFEF}0.8056} \\ \hline
\multicolumn{1}{|l|}{Electricity (num)} &
  \multicolumn{1}{l|}{0.8300} &
  \multicolumn{1}{l|}{0.8320} &
  \multicolumn{1}{l|}{0.8316} &
  \multicolumn{1}{l|}{0.8318} &
  \multicolumn{1}{l|}{0.8618} &
  \multicolumn{1}{l|}{0.8655} &
  \multicolumn{1}{l|}{0.8603} &
  \multicolumn{1}{l|}{0.8629} &
  \multicolumn{1}{l|}{0.7592} &
  \multicolumn{1}{l|}{0.7629} &
  \multicolumn{1}{l|}{0.7596} &
  \multicolumn{1}{l|}{0.7613} \\ \hline
\rowcolor[HTML]{EFEFEF} 
\multicolumn{1}{|l|}{\cellcolor[HTML]{EFEFEF}Adult (num)} &
  \multicolumn{1}{l|}{\cellcolor[HTML]{EFEFEF}0.8197} &
  \multicolumn{1}{l|}{\cellcolor[HTML]{EFEFEF}0.6361} &
  \multicolumn{1}{l|}{\cellcolor[HTML]{EFEFEF}0.6072} &
  \multicolumn{1}{l|}{\cellcolor[HTML]{EFEFEF}0.6213} &
  \multicolumn{1}{l|}{\cellcolor[HTML]{EFEFEF}0.8469} &
  \multicolumn{1}{l|}{\cellcolor[HTML]{EFEFEF}0.7117} &
  \multicolumn{1}{l|}{\cellcolor[HTML]{EFEFEF}0.6242} &
  \multicolumn{1}{l|}{\cellcolor[HTML]{EFEFEF}0.6651} &
  \multicolumn{1}{l|}{\cellcolor[HTML]{EFEFEF}0.8506} &
  \multicolumn{1}{l|}{\cellcolor[HTML]{EFEFEF}0.7600} &
  \multicolumn{1}{l|}{\cellcolor[HTML]{EFEFEF}0.5649} &
  \multicolumn{1}{l|}{\cellcolor[HTML]{EFEFEF}0.6481} \\ \hline
\multicolumn{1}{|l|}{German (num)} &
  \multicolumn{1}{l|}{0.6500} &
  \multicolumn{1}{l|}{0.4286} &
  \multicolumn{1}{l|}{0.4426} &
  \multicolumn{1}{l|}{0.4355} &
  \multicolumn{1}{l|}{0.7700} &
  \multicolumn{1}{l|}{0.6829} &
  \multicolumn{1}{l|}{0.4590} &
  \multicolumn{1}{l|}{0.5490} &
  \multicolumn{1}{l|}{0.7650} &
  \multicolumn{1}{l|}{0.6522} &
  \multicolumn{1}{l|}{0.4918} &
  \multicolumn{1}{l|}{0.5607} \\ \hline
\rowcolor[HTML]{EFEFEF} 
\multicolumn{1}{|l|}{\cellcolor[HTML]{EFEFEF}COMPAS (num)} &
  \multicolumn{1}{l|}{\cellcolor[HTML]{EFEFEF}0.7387} &
  \multicolumn{1}{l|}{\cellcolor[HTML]{EFEFEF}0.8480} &
  \multicolumn{1}{l|}{\cellcolor[HTML]{EFEFEF}0.7974} &
  \multicolumn{1}{l|}{\cellcolor[HTML]{EFEFEF}0.8219} &
  \multicolumn{1}{l|}{\cellcolor[HTML]{EFEFEF}0.7893} &
  \multicolumn{1}{l|}{\cellcolor[HTML]{EFEFEF}0.8529} &
  \multicolumn{1}{l|}{\cellcolor[HTML]{EFEFEF}0.8717} &
  \multicolumn{1}{l|}{\cellcolor[HTML]{EFEFEF}0.8622} &
  \multicolumn{1}{l|}{\cellcolor[HTML]{EFEFEF}0.8170} &
  \multicolumn{1}{l|}{\cellcolor[HTML]{EFEFEF}0.8783} &
  \multicolumn{1}{l|}{\cellcolor[HTML]{EFEFEF}0.8799} &
  \multicolumn{1}{l|}{\cellcolor[HTML]{EFEFEF}0.8791} \\ \hline
\multicolumn{1}{|l|}{Pima Diabetes (num)} &
  \multicolumn{1}{l|}{0.7662} &
  \multicolumn{1}{l|}{0.6719} &
  \multicolumn{1}{l|}{0.7414} &
  \multicolumn{1}{l|}{0.7049} &
  \multicolumn{1}{l|}{0.7727} &
  \multicolumn{1}{l|}{0.7170} &
  \multicolumn{1}{l|}{0.6552} &
  \multicolumn{1}{l|}{0.6847} &
  \multicolumn{1}{l|}{0.7662} &
  \multicolumn{1}{l|}{0.8235} &
  \multicolumn{1}{l|}{0.4828} &
  \multicolumn{1}{l|}{0.6087} \\ \hline
\rowcolor[HTML]{EFEFEF} 
\multicolumn{1}{|l|}{\cellcolor[HTML]{EFEFEF}Breast Cancer (num)} &
  \multicolumn{1}{l|}{\cellcolor[HTML]{EFEFEF}0.9737} &
  \multicolumn{1}{l|}{\cellcolor[HTML]{EFEFEF}0.9750} &
  \multicolumn{1}{l|}{\cellcolor[HTML]{EFEFEF}0.9512} &
  \multicolumn{1}{l|}{\cellcolor[HTML]{EFEFEF}0.9630} &
  \multicolumn{1}{l|}{\cellcolor[HTML]{EFEFEF}0.9912} &
  \multicolumn{1}{l|}{\cellcolor[HTML]{EFEFEF}1.0000} &
  \multicolumn{1}{l|}{\cellcolor[HTML]{EFEFEF}0.9756} &
  \multicolumn{1}{l|}{\cellcolor[HTML]{EFEFEF}0.9877} &
  \multicolumn{1}{l|}{\cellcolor[HTML]{EFEFEF}0.9737} &
  \multicolumn{1}{l|}{\cellcolor[HTML]{EFEFEF}1.0000} &
  \multicolumn{1}{l|}{\cellcolor[HTML]{EFEFEF}0.9268} &
  \multicolumn{1}{l|}{\cellcolor[HTML]{EFEFEF}0.9620} \\ \hline
 
\end{tabular}
}
\end{table}

\section{Detailed Results for Numerical Datasets}

In this section, we present the detailed results for various counterfactual algorithms applied to multiple numerical datasets (Table \ref{tab:num_data}) and mixed variable datasets (Table \ref{tab:mixed_data}) for different machine learning algorithms. The accompanying tables comprehensively compare the performance metrics for each algorithm across these datasets.


{
\tiny
\begin{longtable}{|l|l|l|llll|lllll|lll|}
\caption{Experimental results for numerical datasets:IMAD-$L_1$ refers to median absolute deviation; $MD$ refers to the Mahalanobis Distance; $Spa$ refers to sparsity; $Pla$ refers to plausibility; $Fea$ refers to feasibility; $Div$ refers to diversity; $Cov$ refers to coverage; $Sta$ refers to stability; and $Eff$ refers to efficiency}.
\label{tab:num_data} \\
\hline
\multicolumn{1}{|c|}{\multirow{2}{*}{\textbf{Dataset}}} &
  \multirow{2}{*}{\textbf{\begin{tabular}[c]{@{}l@{}}ML\\Model\end{tabular}}} &
  \multirow{2}{*}{\textbf{Counterfactual}} &
  \multicolumn{4}{c|}{\textbf{Proximity}} &
  \multicolumn{5}{c|}{\textbf{Interpretability}} &
  \multicolumn{3}{c|}{\textbf{Functionality}} \\ \cline{4-15} 
\multicolumn{1}{|c|}{} &
   &
   &
  \multicolumn{1}{l|}{\textbf{L1}} &
  \multicolumn{1}{l|}{\textbf{L2}} &
  \multicolumn{1}{l|}{\textbf{IMAD}} &
  \textbf{MD} &
  \multicolumn{1}{l|}{\textbf{Spa}} &
  \multicolumn{1}{l|}{\textbf{SpaR}} &
  \multicolumn{1}{l|}{\textbf{Pla}} &
  \multicolumn{1}{l|}{\textbf{Fea}} &
  \textbf{Div} &
  \multicolumn{1}{l|}{\textbf{Sta}} &
  \multicolumn{1}{l|}{\textbf{Cov}} &
  \textbf{Eff} \\ 
  \hline
\endfirsthead

\endhead
\multirow{12}{*}{\begin{tabular}[c]{@{}c@{}}\textbf{Breast}\\ \textbf{Cancer} \\ NumFeat: 30  \end{tabular}} &
  \multirow{4}{*}{\textbf{DT}} &
  \textbf{DiCE} &
  \multicolumn{1}{l|}{0.94} &
  \multicolumn{1}{l|}{0.68} &
  \multicolumn{1}{l|}{0.37} &
  0.100 &
  \multicolumn{1}{l|}{2.43} &
  \multicolumn{1}{l|}{0.08} &
  \multicolumn{1}{l|}{\greencheck} &
  \multicolumn{1}{l|}{\greencheck} &
   \greencheck & 
  \multicolumn{1}{l|}{\crosscheck} &
  \multicolumn{1}{l|}{\cellcolor{myBlue}{\textbf{1.00}}} &
  0.03 \\ \cline{3-15} 
 &
   &
  \textbf{GrowingSpheresCF} &
  \multicolumn{1}{l|}{\cellcolor{myBlue}{\textbf{0.13}}} &
  \multicolumn{1}{l|}{\cellcolor{myBlue}{\textbf{0.11}}} &
  \multicolumn{1}{l|}{\cellcolor{myBlue}{\textbf{0.04}}} &
  \cellcolor{myBlue}{\textbf{0.02}} &
  \multicolumn{1}{l|}{\cellcolor{myBlue}{\textbf{1.73}}} &
  \multicolumn{1}{l|}{\cellcolor{myBlue}{\textbf{0.06}}} &
  \multicolumn{1}{l|}{\crosscheck} &
  \multicolumn{1}{l|}{\crosscheck} &
   \crosscheck &
  \multicolumn{1}{l|}{\crosscheck} &
  \multicolumn{1}{l|}{\cellcolor{myBlue}{\textbf{1.00}}} &
  0.03 \\ \cline{3-15} 
 &
   &
  \textbf{Prototype} &
  \multicolumn{1}{l|}{9.83} &
  \multicolumn{1}{l|}{2.06} &
  \multicolumn{1}{l|}{2.98} &
  1.05 &
  \multicolumn{1}{l|}{29.60} &
  \multicolumn{1}{l|}{0.99} &
  \multicolumn{1}{l|}{\crosscheck} &
  \multicolumn{1}{l|}{\crosscheck} &
   \crosscheck &
  \multicolumn{1}{l|}{\greencheck} &
  \multicolumn{1}{l|}{\cellcolor{myOrange}{\textbf{0.25}}} &
  24.72 \\
\hline 

\end{longtable}
}

\begin{table}[!h]
\caption{Experimental results for mixed datasets: \textit{IMAD-L$_1$} refers to median absolute deviation; \textit{MD} refers to the Mahalanobis Distance; \textit{Spa} refers to sparsity; \textit{Pla} refers to plausibility; \textit{Fea} refers to feasibility; \textit{Div} refers to diversity; \textit{Cov} refers to coverage; \textit{Sta} refers to stability; and \textit{Eff} refers to efficiency}
\label{tab:mixed_data}
  \centering\resizebox{\columnwidth}{!} {
%
}}  
& \multicolumn{1}{c|}{{\textbf{Prototype}}}                                 
& \multicolumn{1}{c|}{\cellcolor{myOrange}{12.88}} 
& \multicolumn{1}{c|}{\cellcolor{myOrange}{3.51}}  
& \multicolumn{1}{c|}{2.88}             
& \multicolumn{1}{c?}{\cellcolor{myOrange}{{1.57}}} 
& \multicolumn{1}{c|}{\cellcolor{myOrange}{{8.40}}}            
& \multicolumn{1}{c|}{\cellcolor{myOrange}{{0.70}}}           
& \multicolumn{1}{c|}{{\crosscheck}}    
& \multicolumn{1}{c|}{{\crosscheck}}    
& \multicolumn{1}{c?}{{\crosscheck}}    
& \multicolumn{1}{c|}{{\greencheck}}    
& \multicolumn{1}{c|}{\cellcolor{myOrange}{{0.25}}}           

& \multicolumn{1}{c|}{{29.08}}  \\      
\cline{2-15}

 \multicolumn{1}{|c|}{{}}                              
& \multicolumn{1}{c|}{}                                 
& \multicolumn{1}{c|}{{{DiCE}}}                           
& \multicolumn{1}{c|}{{\cellcolor{myBlue}{{1.87}}}}           
& \multicolumn{1}{c|}{{\cellcolor{myBlue}{1.20}}}
& \multicolumn{1}{c|}{\cellcolor{myOrange}{{4.66}}} 
& \multicolumn{1}{c?}{\cellcolor{myBlue}{{0.27}}}  
& \multicolumn{1}{c|}{{1.74}}           
& \multicolumn{1}{c|}{{0.15}}           
& \multicolumn{1}{c|}{{\greencheck}}    
& \multicolumn{1}{c|}{{\greencheck}}    
& \multicolumn{1}{c?}{{\greencheck}}    
& \multicolumn{1}{c|}{{\crosscheck}}    
& \multicolumn{1}{c|}{\cellcolor{myBlue}{{1.00}}}            
& \multicolumn{1}{c|}{{0.19}} \\        
\cline{3-15}

\multicolumn{1}{|c|}{{ }}                               
& \multicolumn{1}{c|}{\multirow{-2}{*}{\begin{tabular}[c]{@{}c@{}}\textbf{Random} \\ \textbf{Forest}\end{tabular}}}  
& \multicolumn{1}{c|}{{\textbf{Prototype}}}                      
& \multicolumn{1}{c|}{\cellcolor{myOrange}{{12.88}}} 
& \multicolumn{1}{c|}{\cellcolor{myOrange}{{3.51}}}  
& \multicolumn{1}{c|}{{2.88}}            
& \multicolumn{1}{c?}{\cellcolor{myOrange}{{1.57}}}            
& \multicolumn{1}{c|}{\cellcolor{myOrange}{{8.40}}}            
& \multicolumn{1}{c|}{\cellcolor{myOrange}{{0.70}}}            
& \multicolumn{1}{c|}{{\crosscheck}}     
& \multicolumn{1}{c|}{{\crosscheck}}     
& \multicolumn{1}{c?}{{\crosscheck}}     
& \multicolumn{1}{c|}{{\greencheck}}    
& \multicolumn{1}{c|}{\cellcolor{myOrange}{{0.25}}}            
& \multicolumn{1}{c|}{\cellcolor{myOrange}{{264.75}}} \\       
\cline{2-15}

\multicolumn{1}{|c|}{{}}  
& \multicolumn{1}{c|}{}                                 
& \multicolumn{1}{c|}{{\textbf{DiCE}}}                           
& \multicolumn{1}{c|}{2.01}           
& \multicolumn{1}{c|}{{1.27}}           
& \multicolumn{1}{c|}{{5.74}}           
& \multicolumn{1}{c?}{{0.28}}           
& \multicolumn{1}{c|}{\cellcolor{myBlue}{{1.65}}}  
& \multicolumn{1}{c|}{\cellcolor{myBlue}{{0.14}}}  
& \multicolumn{1}{c|}{{\greencheck}}    
& \multicolumn{1}{c|}{{\greencheck}}    
& \multicolumn{1}{c?}{{\greencheck}}    
& \multicolumn{1}{c|}{{\crosscheck}}    
& \multicolumn{1}{c|}{\cellcolor{myBlue}{{1.00}}}           
& \multicolumn{1}{c|}{{0.20}} \\        
\cline{3-15}

\multicolumn{1}{|c|}{\multirow{-6}{*}{{ \begin{tabular}[c]{@{}c@{}}\textbf{Adult}\\ \textbf{Income} \\ NumFeat: 4 \\ CatFeat: 8\\ EncFeat: 103\end{tabular}}}}
& \multicolumn{1}{c|}{\multirow{-2}{*}{\begin{tabular}[c]{@{}c@{}}\textbf{Neural} \\ \textbf{Network}\end{tabular}}} 
& \multicolumn{1}{c|}{{\textbf{Prototype}}}                     
& \multicolumn{1}{c|}{{12.86}}              
& \multicolumn{1}{c|}{\cellcolor{myOrange}{{3.51}}}     
& \multicolumn{1}{c|}{\cellcolor{myBlue}{2.84}}     
& \multicolumn{1}{c?}{\cellcolor{myOrange}{{1.57}}}     
& \multicolumn{1}{c|}{\cellcolor{myOrange}{{8.40}}}               
& \multicolumn{1}{c|}{\cellcolor{myOrange}{{0.70}}}     
& \multicolumn{1}{c|}{{\crosscheck}}        
& \multicolumn{1}{c|}{{\crosscheck}}        
& \multicolumn{1}{c?}{{\crosscheck}}        
& \multicolumn{1}{c|}{\greencheck}          
& \multicolumn{1}{c|}{\cellcolor{myOrange}{{0.25}}}               
& \multicolumn{1}{c|}{{82.46}}  \\          
\hline
\hline

\multicolumn{1}{|c|}{{ }}                               
& \multicolumn{1}{c|}{}                             
& \multicolumn{1}{c|}{{\textbf{DiCE}}}                           
& \multicolumn{1}{c|}{{1.19}}   
& \multicolumn{1}{c|}{{0.88}}                
& \multicolumn{1}{c|}{{3.28}}                
& \multicolumn{1}{c?}{{0.33}}                
& \multicolumn{1}{c|}{\cellcolor{myBlue}{{1.37}}}                
& \multicolumn{1}{c|}{\cellcolor{myBlue}{{0.12}}}                
& \multicolumn{1}{c|}{{\greencheck}}         
& \multicolumn{1}{c|}{{\greencheck}}         
& \multicolumn{1}{c?}{{\greencheck}}         
& \multicolumn{1}{c|}{{\crosscheck}}         
& \multicolumn{1}{c|}{\cellcolor{myBlue}{\textbf{1.00}}}                
& \multicolumn{1}{c|}{\cellcolor{myBlue}{{0.06}}}    \\          
\cline{3-15}

\multicolumn{1}{|c|}{{}}                                
& \multicolumn{1}{c|}{\multirow{-2}{*}{\begin{tabular}[c]{@{}c@{}}\textbf{Decision} \\ \textbf{Tree}\end{tabular}}}  
& \multicolumn{1}{c|}{{\textbf{Prototype}}}                      
& \multicolumn{1}{c|}{{7.07}}               
& \multicolumn{1}{c|}{{2.58}}               
& \multicolumn{1}{c|}{\cellcolor{myOrange}{{3.96}}}  
& \multicolumn{1}{c?}{{1.77}}               
& \multicolumn{1}{c|}{{6.50}}               
& \multicolumn{1}{c|}{{0.59}}               
& \multicolumn{1}{c|}{{\crosscheck}}        
& \multicolumn{1}{c|}{{\crosscheck}}        
& \multicolumn{1}{c?}{{\crosscheck}}        
& \multicolumn{1}{c|}{{\greencheck}}        
& \multicolumn{1}{c|}{{0.20}}                
& \multicolumn{1}{c|}{{32.10}} \\           
\cline{2-15}

\multicolumn{1}{|c|}{{ }}                               
& \multicolumn{1}{c|}{}                             
& \multicolumn{1}{c|}{{\textbf{DiCE}}}                           
& \multicolumn{1}{c|}{\cellcolor{myBlue}{{1.16}}}               
& \multicolumn{1}{c|}{\cellcolor{myBlue}{{0.85}}}               
& \multicolumn{1}{c|}{{3.71}}     
& \multicolumn{1}{c?}{\cellcolor{myBlue}{{0.25}}}               
& \multicolumn{1}{c|}{{1.52}}               
& \multicolumn{1}{c|}{{0.14}}               
& \multicolumn{1}{c|}{{\greencheck}}        
& \multicolumn{1}{c|}{{\greencheck}}        
& \multicolumn{1}{c?}{{ \greencheck}}       
& \multicolumn{1}{c|}{{\crosscheck}}        
& \multicolumn{1}{c|}{\cellcolor{myBlue}{{1.00}}}               
& \multicolumn{1}{c|}{{0.09}}   \\          
\cline{3-15}

\multicolumn{1}{|c|}{{ }}   
& \multicolumn{1}{c|}{\multirow{-2}{*}{\begin{tabular}[c]{@{}c@{}}\textbf{Random} \\ \textbf{Forest}\end{tabular}}}  
& \multicolumn{1}{c|}{{\textbf{Prototype}}}      
& \multicolumn{1}{c|}{\cellcolor{myOrange}{{7.51}}}               
& \multicolumn{1}{c|}{\cellcolor{myOrange}{{2.68}}}               
& \multicolumn{1}{c|}{{3.40}}               
& \multicolumn{1}{c?}{\cellcolor{myOrange}{{1.98}}}               
& \multicolumn{1}{c|}{\cellcolor{myOrange}{{7.00}}}               
& \multicolumn{1}{c|}{\cellcolor{myOrange}{{0.64}}}               
& \multicolumn{1}{c|}{{\crosscheck}}        
& \multicolumn{1}{c|}{{\crosscheck}}        
& \multicolumn{1}{c?}{{\crosscheck}}        
& \multicolumn{1}{c|}{{\greencheck}}        
& \multicolumn{1}{c|}{{0.20}}               
& \multicolumn{1}{c|}{\cellcolor{myOrange}{{290.92}}} \\          
\cline{2-15}

\multicolumn{1}{|c|}{{ }}               
& \multicolumn{1}{c|}{}                                 
& \multicolumn{1}{c|}{{\textbf{DiCE}}}                           
& \multicolumn{1}{c|}{{1.45}}               
& \multicolumn{1}{c|}{{1.02}}               
& \multicolumn{1}{c|}{{4.59}}               
& \multicolumn{1}{c?}{{0.36}}               
& \multicolumn{1}{c|}{{1.60}}               
& \multicolumn{1}{c|}{{0.15}}               
& \multicolumn{1}{c|}{{\greencheck}}        
& \multicolumn{1}{c|}{{\greencheck}}        
& \multicolumn{1}{c?}{{\greencheck}}        
& \multicolumn{1}{c|}{{\crosscheck}}        
& \multicolumn{1}{c|}{\cellcolor{myBlue}{{1.00}}}               
& \multicolumn{1}{c|}{{0.10}}   \\          
\cline{3-15}

\multicolumn{1}{|c|}{\multirow{-6}{*}{{  \begin{tabular}[c]{@{}c@{}}\textbf{COMPAS}\\  NumFeat: 4 \\ CatFeat: 7\\ EncFeat: 23\end{tabular}  }}}    
& \multicolumn{1}{c|}{\multirow{-2}{*}{\begin{tabular}[c]{@{}c@{}}\textbf{Neural} \\ \textbf{Network}\end{tabular}}} 
& \multicolumn{1}{c|}{{\textbf{Prototype}}}                     
& \multicolumn{1}{c|}{{5.48}}               
& \multicolumn{1}{c|}{{2.24}}               
& \multicolumn{1}{c|}{\cellcolor{myBlue}{{3.20}}}               
& \multicolumn{1}{c?}{{1.31}}               
& \multicolumn{1}{c|}{{6.00}}               
& \multicolumn{1}{c|}{{0.55}}               
& \multicolumn{1}{c|}{{\crosscheck}}        
& \multicolumn{1}{c|}{{\crosscheck}}        
& \multicolumn{1}{c?}{{\crosscheck}}        
& \multicolumn{1}{c|}{{\greencheck}}        
& \multicolumn{1}{c|}{\cellcolor{myOrange}{{0.10}}}               
& \multicolumn{1}{c|}{{76.20}}   \\         
\hline
\hline

\multicolumn{1}{|c|}{{ }}                   
& \multicolumn{1}{c|}{}     
& \multicolumn{1}{c|}{{\textbf{DiCE}}}                           
& \multicolumn{1}{c|}{\cellcolor{myBlue}{{2.52}}}               
& \multicolumn{1}{c|}{\cellcolor{myBlue}{{1.44}}}               
& \multicolumn{1}{c|}{\cellcolor{myBlue}{{0.37}}}               
& \multicolumn{1}{c?}{\cellcolor{myBlue}{{0.60}}}               
& \multicolumn{1}{c|}{\cellcolor{myBlue}{{1.68}}}               
& \multicolumn{1}{c|}{{\cellcolor{myBlue}{0.08}}}               
& \multicolumn{1}{c|}{{\greencheck}}        
& \multicolumn{1}{c|}{{\greencheck}}        
& \multicolumn{1}{c?}{{\greencheck}}        
& \multicolumn{1}{c|}{{\crosscheck}}        
& \multicolumn{1}{c|}{\cellcolor{myBlue}{{1.00}}}               
& \multicolumn{1}{c|}{\cellcolor{myBlue}{{0.14}}}   \\          
\cline{3-15}

\multicolumn{1}{|c|}{{ }}      
& \multicolumn{1}{c|}{\multirow{-2}{*}{\begin{tabular}[c]{@{}c@{}}\textbf{Decision} \\ \textbf{Tree}\end{tabular}}}  
& \multicolumn{1}{c|}{{{Prototype}}}                      
& \multicolumn{1}{c|}{{21.32}}              
& \multicolumn{1}{c|}{{4.55}}               
& \multicolumn{1}{c|}{{1.30}}               
& \multicolumn{1}{c?}{{2.34}}               
& \multicolumn{1}{c|}{{14.00}}              
& \multicolumn{1}{c|}{{0.70}}               
& \multicolumn{1}{c|}{{\crosscheck}}        
& \multicolumn{1}{c|}{{\crosscheck}}        
& \multicolumn{1}{c?}{{\crosscheck}}        
& \multicolumn{1}{c|}{{\greencheck}}        
& \multicolumn{1}{c|}{{0.50}}               
& \multicolumn{1}{c|}{{30.39}}  \\          
\cline{2-15}

\multicolumn{1}{|c|}{{ }}   
& \multicolumn{1}{c|}{}                                 
& \multicolumn{1}{c|}{{\textbf{DiCE}}}                           
& \multicolumn{1}{c|}{{3.67}}               
& \multicolumn{1}{c|}{{1.73}}               
& \multicolumn{1}{c|}{{0.54}}               
& \multicolumn{1}{c?}{{0.80}}               
& \multicolumn{1}{c|}{{2.36}}               
& \multicolumn{1}{c|}{{0.19}}               
& \multicolumn{1}{c|}{{\greencheck}}        
& \multicolumn{1}{c|}{{\greencheck}}        
& \multicolumn{1}{c?}{{\greencheck}}        
& \multicolumn{1}{c|}{{\crosscheck}}        
& \multicolumn{1}{c|}{\cellcolor{myBlue}{{1.00}}}               
& \multicolumn{1}{c|}{{0.20}}  \\           
\cline{3-15}

\multicolumn{1}{|c|}{{ }}   
& \multicolumn{1}{c|}{\multirow{-2}{*}{\begin{tabular}[c]{@{}c@{}}\textbf{Random} \\ \textbf{Forest}\end{tabular}}}  
& \multicolumn{1}{c|}{{\textbf{Prototype}}}                  
& \multicolumn{1}{c|}{\cellcolor{myOrange}{{23.20}}} 
& \multicolumn{1}{c|}{\cellcolor{myOrange}{{4.74}}}  
& \multicolumn{1}{c|}{\cellcolor{myOrange}{{1.95}}}  
& \multicolumn{1}{c?}{{2.47}}               
& \multicolumn{1}{c|}{\cellcolor{myOrange}{{14.63}}}              
& \multicolumn{1}{c|}{\cellcolor{myOrange}{{0.73}}}               
& \multicolumn{1}{c|}{{\crosscheck}}        
& \multicolumn{1}{c|}{{\crosscheck}}        
& \multicolumn{1}{c?}{{\crosscheck}}        
& \multicolumn{1}{c|}{{\greencheck}}        
& \multicolumn{1}{c|}{\cellcolor{myOrange}{{0.40}}} 
& \multicolumn{1}{c|}{\cellcolor{myOrange}{{249.68}}} \\          

\cline{2-15} 
\multicolumn{1}{|c|}{{ }}                               
& \multicolumn{1}{c|}{}                                 
& \multicolumn{1}{c|}{{\textbf{DiCE}}}                           
& \multicolumn{1}{c|}{{3.79}}               
& \multicolumn{1}{c|}{{1.81}}               
& \multicolumn{1}{c|}{{0.47}}               
& \multicolumn{1}{c?}{{0.74}}               
& \multicolumn{1}{c|}{{2.36}}               
& \multicolumn{1}{c|}{{0.12}}               
& \multicolumn{1}{c|}{{\greencheck}}        
& \multicolumn{1}{c|}{{\greencheck}}        
& \multicolumn{1}{c?}{{\greencheck}}        
& \multicolumn{1}{c|}{{\crosscheck}}        
& \multicolumn{1}{c|}{\cellcolor{myBlue}{{1.00}}}               
& \multicolumn{1}{c|}{{0.20}}   \\          
\cline{3-15}

\multicolumn{1}{|c|}{\multirow{-6}{*}{{ \begin{tabular}[c]{@{}c@{}}\textbf{German}\\ \textbf{Credit} \\ NumFeat: 5 \\ CatFeat: 15\\ EncFeat: 65\end{tabular}}}} 
& \multicolumn{1}{c|}{\multirow{-2}{*}{\begin{tabular}[c]{@{}c@{}}\textbf{Neural} \\ \textbf{Network}\end{tabular}}} 
& \multicolumn{1}{c|}{{\textbf{Prototype}}}                      
& \multicolumn{1}{c|}{{22.95}}              
& \multicolumn{1}{c|}{{4.71}}               
& \multicolumn{1}{c|}{{1.83}}               
& \multicolumn{1}{c?}{\cellcolor{myOrange}{{2.49}}}               
& \multicolumn{1}{c|}{{14.60}}              
& \multicolumn{1}{c|}{\cellcolor{myOrange}{{0.73}}}               
& \multicolumn{1}{c|}{{\crosscheck}}        
& \multicolumn{1}{c|}{{\crosscheck}}        
& \multicolumn{1}{c?}{{\crosscheck}}        
& \multicolumn{1}{c|}{{\greencheck}}        
& \multicolumn{1}{c|}{{0.50}}               
& \multicolumn{1}{c|}{{84.73}}  \\          
\hline
\hline

\multicolumn{1}{|c|}{{ }}                   
& \multicolumn{1}{c|}{}     
& \multicolumn{1}{c|}{{\textbf{DiCE}}}                           
& \multicolumn{1}{c|}{\cellcolor{myBlue}{{0.77}}}               
& \multicolumn{1}{c|}{\cellcolor{myBlue}{{0.63}}}               
& \multicolumn{1}{c|}{2.68}            
& \multicolumn{1}{c?}{\cellcolor{myBlue}{{0.04}}}               
& \multicolumn{1}{c|}{\cellcolor{myBlue}{{1.54}}}               
& \multicolumn{1}{c|}{{\cellcolor{myBlue}{{0.05}}}}               
& \multicolumn{1}{c|}{{\greencheck}}        
& \multicolumn{1}{c|}{{\greencheck}}        
& \multicolumn{1}{c?}{{\greencheck}}        
& \multicolumn{1}{c|}{{\crosscheck}}        
& \multicolumn{1}{c|}{\cellcolor{myBlue}{{0.65}}}               
& \multicolumn{1}{c|}{\cellcolor{myBlue}{{0.04}}}   \\          
\cline{3-15}

\multicolumn{1}{|c|}{{ }}      
& \multicolumn{1}{c|}{\multirow{-2}{*}{\begin{tabular}[c]{@{}c@{}}\textbf{Decision} \\ \textbf{Tree}\end{tabular}}}  
& \multicolumn{1}{c|}{{\textbf{Prototype}}}                      
& \multicolumn{1}{c|}{{\cellcolor{myOrange}3.64}}              
& \multicolumn{1}{c|}{{\cellcolor{myOrange}1.37}}               
& \multicolumn{1}{c|}{{1.12}}               
& \multicolumn{1}{c?}{{0.34}}               
& \multicolumn{1}{c|}{{23.89}}              
& \multicolumn{1}{c|}{{0.75}}               
& \multicolumn{1}{c|}{{\crosscheck}}        
& \multicolumn{1}{c|}{{\crosscheck}}        
& \multicolumn{1}{c?}{{\crosscheck}}        
& \multicolumn{1}{c|}{{\greencheck}}        
& \multicolumn{1}{c|}{{0.09}}               
& \multicolumn{1}{c|}{{23.73}}  \\          
\cline{2-15}

\multicolumn{1}{|c|}{{ }}   
& \multicolumn{1}{c|}{}                                 
& \multicolumn{1}{c|}{{\textbf{DiCE}}}                           
& \multicolumn{1}{c|}{{1.12}}               
& \multicolumn{1}{c|}{{0.82}}               
& \multicolumn{1}{c|}{{5.70}}               
& \multicolumn{1}{c?}{{0.05}}               
& \multicolumn{1}{c|}{{2.13}}               
& \multicolumn{1}{c|}{{0.07}}               
& \multicolumn{1}{c|}{{\greencheck}}        
& \multicolumn{1}{c|}{{\greencheck}}        
& \multicolumn{1}{c?}{{\greencheck}}        
& \multicolumn{1}{c|}{{\crosscheck}}        
& \multicolumn{1}{c|}{0.75}                 
& \multicolumn{1}{c|}{{0.05}}  \\           
\cline{3-15}

\multicolumn{1}{|c|}{\multirow{-3}{*}{{  \begin{tabular}[c]{@{}c@{}}\textbf{Albert}\\  NumFeat: 21 \\ CatFeat: 10\\ EncFeat: 10\end{tabular}  }}}
& \multicolumn{1}{c|}{\multirow{-2}{*}{\begin{tabular}[c]{@{}c@{}}\textbf{Random} \\ \textbf{Forest}\end{tabular}}}  
& \multicolumn{1}{c|}{{\textbf{Prototype}}}                  
& \multicolumn{1}{c|}{{3.5}} 
& \multicolumn{1}{c|}{{1.34}}  
& \multicolumn{1}{c|}{{1.05}}  
& \multicolumn{1}{c?}{{0.33}}               
& \multicolumn{1}{c|}{{24.09}}              
& \multicolumn{1}{c|}{{0.75}}               
& \multicolumn{1}{c|}{{\crosscheck}}        
& \multicolumn{1}{c|}{{\crosscheck}}        
& \multicolumn{1}{c?}{{\crosscheck}}        
& \multicolumn{1}{c|}{{\greencheck}}        
& \multicolumn{1}{c|}{{0.11}} 
& \multicolumn{1}{c|}{{124.83}} \\          

\cline{2-15} 
\multicolumn{1}{|c|}{{ }}                               
& \multicolumn{1}{c|}{}                                 
& \multicolumn{1}{c|}{{\textbf{DiCE}}}                           
& \multicolumn{1}{c|}{{0.86}}               
& \multicolumn{1}{c|}{{0.70}}               
& \multicolumn{1}{c|}{{2.85}}               
& \multicolumn{1}{c?}{{0.06}}               
& \multicolumn{1}{c|}{{1.62}}               
& \multicolumn{1}{c|}{{\cellcolor{myBlue}0.05}}               
& \multicolumn{1}{c|}{{\greencheck}}        
& \multicolumn{1}{c|}{{\greencheck}}        
& \multicolumn{1}{c?}{{\greencheck}}        
& \multicolumn{1}{c|}{{\crosscheck}}        
& \multicolumn{1}{c|}{\cellcolor{myBlue}{{0.65}}}               
& \multicolumn{1}{c|}{{0.04}}   \\          
\cline{3-15}

\multicolumn{1}{|c|}{\multirow{-6}{*}} & \multicolumn{1}{c|}{\multirow{-2}{*}{\begin{tabular}[c]{@{}c@{}}\textbf{Neural} \\ \textbf{Network}\end{tabular}}} 
& \multicolumn{1}{c|}{{\textbf{Prototype}}}                      
& \multicolumn{1}{c|}{{3.51}}              
& \multicolumn{1}{c|}{{1.26}}               
& \multicolumn{1}{c|}{{\cellcolor{myOrange}1.22}}               
& \multicolumn{1}{c?}{0.30}               
& \multicolumn{1}{c|}{{23.60}}              
& \multicolumn{1}{c|}{\cellcolor{myOrange}{{0.74}}}               
& \multicolumn{1}{c|}{{\crosscheck}}        
& \multicolumn{1}{c|}{{\crosscheck}}        
& \multicolumn{1}{c?}{{\crosscheck}}        
& \multicolumn{1}{c|}{{\greencheck}}        
& \multicolumn{1}{c|}{{0.10}}               
& \multicolumn{1}{c|}{{29.77}}  \\          
\hline
\hline

\multicolumn{1}{|c|}{{ }}                   
& \multicolumn{1}{c|}{}     
& \multicolumn{1}{c|}{{\textbf{DiCE}}}                           
& \multicolumn{1}{c|}{\cellcolor{myBlue}{{0.62}}}               
& \multicolumn{1}{c|}{\cellcolor{myBlue}{{0.51}}}               
& \multicolumn{1}{c|}{\cellcolor{myBlue}{{2.77}}}               
& \multicolumn{1}{c?}{\cellcolor{myBlue}{{0.10}}}               
& \multicolumn{1}{c|}{\cellcolor{myBlue}{{1.65}}}               
& \multicolumn{1}{c|}{{\cellcolor{myBlue}{0.18}}}               
& \multicolumn{1}{c|}{{\greencheck}}        
& \multicolumn{1}{c|}{{\greencheck}}        
& \multicolumn{1}{c?}{{\greencheck}}        
& \multicolumn{1}{c|}{{\crosscheck}}        
& \multicolumn{1}{c|}{\cellcolor{myBlue}{{0.85}}}               
& \multicolumn{1}{c|}{\cellcolor{myBlue}{{0.02}}}   \\          
\cline{3-15}

\multicolumn{1}{|c|}{{ }}      
& \multicolumn{1}{c|}{\multirow{-2}{*}{\begin{tabular}[c]{@{}c@{}}\textbf{Decision} \\ \textbf{Tree}\end{tabular}}}  
& \multicolumn{1}{c|}{{\textbf{Prototype}}}                      
& \multicolumn{1}{c|}{{2.23}}              
& \multicolumn{1}{c|}{{1.02}}               
& \multicolumn{1}{c|}{{2.72}}               
& \multicolumn{1}{c?}{{0.24}}               
& \multicolumn{1}{c|}{{7.14}}              
& \multicolumn{1}{c|}{{0.79}}               
& \multicolumn{1}{c|}{{\crosscheck}}        
& \multicolumn{1}{c|}{{\crosscheck}}        
& \multicolumn{1}{c?}{{\crosscheck}}        
& \multicolumn{1}{c|}{{\greencheck}}        
& \multicolumn{1}{c|}{{0.35}}               
& \multicolumn{1}{c|}{{24.36}}  \\          
\cline{2-15}

\multicolumn{1}{|c|}{{ }}   
& \multicolumn{1}{c|}{}                                 
& \multicolumn{1}{c|}{{\textbf{DiCE}}}                           
& \multicolumn{1}{c|}{{0.71}}               
& \multicolumn{1}{c|}{{0.56}}               
& \multicolumn{1}{c|}{{12.84}}               
& \multicolumn{1}{c?}{{0.08}}               
& \multicolumn{1}{c|}{{1.82}}               
& \multicolumn{1}{c|}{{0.2}}               
& \multicolumn{1}{c|}{{\greencheck}}        
& \multicolumn{1}{c|}{{\greencheck}}        
& \multicolumn{1}{c?}{{\greencheck}}        
& \multicolumn{1}{c|}{{\crosscheck}}        
& \multicolumn{1}{c|}{\cellcolor{myBlue}{{0.85}}}               
& \multicolumn{1}{c|}{{0.04}}  \\           
\cline{3-15}

\multicolumn{1}{|c|}{{ }}   
& \multicolumn{1}{c|}{\multirow{-2}{*}{\begin{tabular}[c]{@{}c@{}}\textbf{Random} \\ \textbf{Forest}\end{tabular}}}  
& \multicolumn{1}{c|}{{\textbf{Prototype}}}                  
& \multicolumn{1}{c|}{\cellcolor{myOrange}{{2.41}}} 
& \multicolumn{1}{c|}{\cellcolor{myOrange}{{1.07}}}  
& \multicolumn{1}{c|}{\cellcolor{myOrange}{{2.85}}}  
& \multicolumn{1}{c?}{{0.26}}               
& \multicolumn{1}{c|}{\cellcolor{myOrange}{{7.29}}}              
& \multicolumn{1}{c|}{\cellcolor{myOrange}{{0.81}}}               
& \multicolumn{1}{c|}{{\crosscheck}}        
& \multicolumn{1}{c|}{{\crosscheck}}        
& \multicolumn{1}{c?}{{\crosscheck}}        
& \multicolumn{1}{c|}{{\greencheck}}        
& \multicolumn{1}{c|}{\cellcolor{myOrange}{{0.35}}} 
& \multicolumn{1}{c|}{\cellcolor{myOrange}{{126.68}}} \\          

\cline{2-15} 
\multicolumn{1}{|c|}{{ }}                               
& \multicolumn{1}{c|}{}                                 
& \multicolumn{1}{c|}{{\textbf{DiCE}}}                           
& \multicolumn{1}{c|}{{0.67}}               
& \multicolumn{1}{c|}{{0.57}}               
& \multicolumn{1}{c|}{{24.86}}               
& \multicolumn{1}{c?}{{0.06}}               
& \multicolumn{1}{c|}{{1.56}}               
& \multicolumn{1}{c|}{{0.17}}               
& \multicolumn{1}{c|}{{\greencheck}}        
& \multicolumn{1}{c|}{{\greencheck}}        
& \multicolumn{1}{c?}{{\greencheck}}        
& \multicolumn{1}{c|}{{\crosscheck}}        
& \multicolumn{1}{c|}{\cellcolor{myBlue}{{0.8}}}               
& \multicolumn{1}{c|}{{0.03}}   \\          
\cline{3-15} 

\multicolumn{1}{|c|}{\multirow{-6}{*}{ \begin{tabular}[c]{@{}c@{}}\textbf{Electricity}\\  NumFeat: 7 \\ CatFeat: 1\\ EncFeat: 1\end{tabular}} }
& \multicolumn{1}{c|}{\multirow{-2}{*}{\begin{tabular}[c]{@{}c@{}}\textbf{Neural} \\ \textbf{Network}\end{tabular}}} 
& \multicolumn{1}{c|}{{\textbf{Prototype}}}                      
& \multicolumn{1}{c|}{{2.36}}              
& \multicolumn{1}{c|}{{1.08}}               
& \multicolumn{1}{c|}{{2.84}}               
& \multicolumn{1}{c?}{\cellcolor{myOrange}{{0.26}}}               
& \multicolumn{1}{c|}{{7.12}}              
& \multicolumn{1}{c|}{\cellcolor{myOrange}{{0.79}}}               
& \multicolumn{1}{c|}{{\crosscheck}}        
& \multicolumn{1}{c|}{{\crosscheck}}        
& \multicolumn{1}{c?}{{\crosscheck}}        
& \multicolumn{1}{c|}{{\greencheck}}        
& \multicolumn{1}{c|}{{0.40}}               
& \multicolumn{1}{c|}{{27.3}}  \\          
\hline
\hline

\multicolumn{1}{|c|}{{ }}                   
& \multicolumn{1}{c|}{}     
& \multicolumn{1}{c|}{{\textbf{DiCE}}}                           
& \multicolumn{1}{c|}{\cellcolor{myBlue}{{0.78}}}               
& \multicolumn{1}{c|}{\cellcolor{myBlue}{{0.69}}}               
& \multicolumn{1}{c|}{\cellcolor{myBlue}{{0.72}}}               
& \multicolumn{1}{c?}{\cellcolor{myBlue}{{0.10}}}               
& \multicolumn{1}{c|}{\cellcolor{myBlue}{{1.30}}}               
& \multicolumn{1}{c|}{{\cellcolor{myBlue}{0.05}}}               
& \multicolumn{1}{c|}{{\greencheck}}        
& \multicolumn{1}{c|}{{\greencheck}}        
& \multicolumn{1}{c?}{{\greencheck}}        
& \multicolumn{1}{c|}{{\crosscheck}}        
& \multicolumn{1}{c|}{\cellcolor{myBlue}{{0.50}}}               
& \multicolumn{1}{c|}{\cellcolor{myBlue}{{0.03}}}   \\          
\cline{3-15}

\multicolumn{1}{|c|}{{ }}      
& \multicolumn{1}{c|}{\multirow{-2}{*}{\begin{tabular}[c]{@{}c@{}}\textbf{Decision} \\ \textbf{Tree}\end{tabular}}}  
& \multicolumn{1}{c|}{{\textbf{Prototype}}}                      
& \multicolumn{1}{c|}{{3.87}}              
& \multicolumn{1}{c|}{{1.20}}               
& \multicolumn{1}{c|}{{2.75}}               
& \multicolumn{1}{c?}{{0.32}}               
& \multicolumn{1}{c|}{{18.00}}              
& \multicolumn{1}{c|}{{0.75}}               
& \multicolumn{1}{c|}{{\crosscheck}}        
& \multicolumn{1}{c|}{{\crosscheck}}        
& \multicolumn{1}{c?}{{\crosscheck}}        
& \multicolumn{1}{c|}{{\greencheck}}        
& \multicolumn{1}{c|}{{0.35}}               
& \multicolumn{1}{c|}{{11.14}}  \\          
\cline{2-15}

\multicolumn{1}{|c|}{{ }}   
& \multicolumn{1}{c|}{}                                 
& \multicolumn{1}{c|}{{\textbf{DiCE}}}                           
& \multicolumn{1}{c|}{{0.76}}               
& \multicolumn{1}{c|}{{0.63}}               
& \multicolumn{1}{c|}{{0.76}}               
& \multicolumn{1}{c?}{{0.07}}               
& \multicolumn{1}{c|}{{1.67}}               
& \multicolumn{1}{c|}{{0.07}}               
& \multicolumn{1}{c|}{{\greencheck}}        
& \multicolumn{1}{c|}{{\greencheck}}        
& \multicolumn{1}{c?}{{\greencheck}}        
& \multicolumn{1}{c|}{{\crosscheck}}        
& \multicolumn{1}{c|}{\cellcolor{myBlue}{{0.45}}}               
& \multicolumn{1}{c|}{{0.04}}  \\           
\cline{3-15}

\multicolumn{1}{|c|}{{ }}   
& \multicolumn{1}{c|}{\multirow{-2}{*}{\begin{tabular}[c]{@{}c@{}}\textbf{Random} \\ \textbf{Forest}\end{tabular}}}  
& \multicolumn{1}{c|}{{\textbf{Prototype}}}                  
& \multicolumn{1}{c|}{\cellcolor{myOrange}{{3.55}}} 
& \multicolumn{1}{c|}{\cellcolor{myOrange}{{1.14}}}  
& \multicolumn{1}{c|}{\cellcolor{myOrange}{{2.71}}}  
& \multicolumn{1}{c?}{{0.29}}               
& \multicolumn{1}{c|}{\cellcolor{myOrange}{{18.00}}}              
& \multicolumn{1}{c|}{\cellcolor{myOrange}{{0.75}}}               
& \multicolumn{1}{c|}{{\crosscheck}}        
& \multicolumn{1}{c|}{{\crosscheck}}        
& \multicolumn{1}{c?}{{\crosscheck}}        
& \multicolumn{1}{c|}{{\greencheck}}        
& \multicolumn{1}{c|}{\cellcolor{myOrange}{{0.50}}} 
& \multicolumn{1}{c|}{\cellcolor{myOrange}{{87.83}}} \\          

\cline{2-15} 
\multicolumn{1}{|c|}{{ }}                               
& \multicolumn{1}{c|}{}                                 
& \multicolumn{1}{c|}{{\textbf{DiCE}}}                           
& \multicolumn{1}{c|}{{0.88}}               
& \multicolumn{1}{c|}{{0.71}}               
& \multicolumn{1}{c|}{{0.92}}               
& \multicolumn{1}{c?}{{0.10}}               
& \multicolumn{1}{c|}{{1.67}}               
& \multicolumn{1}{c|}{{0.07}}               
& \multicolumn{1}{c|}{{\greencheck}}        
& \multicolumn{1}{c|}{{\greencheck}}        
& \multicolumn{1}{c?}{{\greencheck}}        
& \multicolumn{1}{c|}{{\crosscheck}}        
& \multicolumn{1}{c|}{\cellcolor{myBlue}{{0.75}}}               
& \multicolumn{1}{c|}{{0.03}}   \\          
\cline{3-15}

\multicolumn{1}{|c|}{\multirow{-6}{*}{{\begin{tabular}[c]{@{}c@{}}\textbf{Eye} \\ \textbf{Movements} \\  NumFeat: 20 \\ CatFeat: 3\\ EncFeat: 3 \end{tabular}}}} 
& \multicolumn{1}{c|}{\multirow{-2}{*}{\begin{tabular}[c]{@{}c@{}}\textbf{Neural} \\ \textbf{Network}\end{tabular}}} 
& \multicolumn{1}{c|}{{\textbf{Prototype}}}                      
& \multicolumn{1}{c|}{{3.79}}              
& \multicolumn{1}{c|}{{1.14}}               
& \multicolumn{1}{c|}{{2.81}}               
& \multicolumn{1}{c?}{\cellcolor{myOrange}{{0.28}}}               
& \multicolumn{1}{c|}{{18.29}}              
& \multicolumn{1}{c|}{\cellcolor{myOrange}{{0.76}}}               
& \multicolumn{1}{c|}{{\crosscheck}}        
& \multicolumn{1}{c|}{{\crosscheck}}        
& \multicolumn{1}{c?}{{\crosscheck}}        
& \multicolumn{1}{c|}{{\greencheck}}        
& \multicolumn{1}{c|}{{0.35}}               
& \multicolumn{1}{c|}{{27.3}}  \\          
\hline
\hline

\multicolumn{1}{|c|}{{ }}                   
& \multicolumn{1}{c|}{}     
& \multicolumn{1}{c|}{{\textbf{DiCE}}}                           
& \multicolumn{1}{c|}{\cellcolor{myBlue}{{0.72}}}               
& \multicolumn{1}{c|}{\cellcolor{myBlue}{{0.59}}}               
& \multicolumn{1}{c|}{\cellcolor{myBlue}{{0.59}}}               
& \multicolumn{1}{c?}{\cellcolor{myBlue}{{0.15}}}               
& \multicolumn{1}{c|}{\cellcolor{myBlue}{{1.59}}}               
& \multicolumn{1}{c|}{{\cellcolor{myBlue}{0.03}}}               
& \multicolumn{1}{c|}{{\greencheck}}        
& \multicolumn{1}{c|}{{\greencheck}}        
& \multicolumn{1}{c?}{{\greencheck}}        
& \multicolumn{1}{c|}{{\crosscheck}}        
& \multicolumn{1}{c|}{\cellcolor{myBlue}{{0.85}}}               
& \multicolumn{1}{c|}{\cellcolor{myBlue}{{0.07}}}   \\          
\cline{3-15}

\multicolumn{1}{|c|}{{ }}      
& \multicolumn{1}{c|}{\multirow{-2}{*}{\begin{tabular}[c]{@{}c@{}}\textbf{Decision} \\ \textbf{Tree}\end{tabular}}}  
& \multicolumn{1}{c|}{{\textbf{Prototype}}}                      
& \multicolumn{1}{c|}{{3.92}}              
& \multicolumn{1}{c|}{{1.50}}               
& \multicolumn{1}{c|}{{0.81}}               
& \multicolumn{1}{c?}{{0.23}}               
& \multicolumn{1}{c|}{{10.00}}              
& \multicolumn{1}{c|}{{0.18}}               
& \multicolumn{1}{c|}{{\crosscheck}}        
& \multicolumn{1}{c|}{{\crosscheck}}        
& \multicolumn{1}{c?}{{\crosscheck}}        
& \multicolumn{1}{c|}{{\greencheck}}        
& \multicolumn{1}{c|}{{0.35}}               
& \multicolumn{1}{c|}{{11.18}}  \\          
\cline{2-15}

\multicolumn{1}{|c|}{{ }}   
& \multicolumn{1}{c|}{}                                 
& \multicolumn{1}{c|}{{\textbf{DiCE}}}                           
& \multicolumn{1}{c|}{{1.11}}               
& \multicolumn{1}{c|}{{0.81}}               
& \multicolumn{1}{c|}{{1.06}}               
& \multicolumn{1}{c?}{{0.15}}               
& \multicolumn{1}{c|}{{1.93}}               
& \multicolumn{1}{c|}{{0.04}}               
& \multicolumn{1}{c|}{{\greencheck}}        
& \multicolumn{1}{c|}{{\greencheck}}        
& \multicolumn{1}{c?}{{\greencheck}}        
& \multicolumn{1}{c|}{{\crosscheck}}        
& \multicolumn{1}{c|}{\cellcolor{myBlue}{{0.75}}}               
& \multicolumn{1}{c|}{{0.08}}  \\           
\cline{3-15}

\multicolumn{1}{|c|}{{ }}   
& \multicolumn{1}{c|}{\multirow{-2}{*}{\begin{tabular}[c]{@{}c@{}}\textbf{Random} \\ \textbf{Forest}\end{tabular}}}  
& \multicolumn{1}{c|}{{\textbf{Prototype}}}                  
& \multicolumn{1}{c|}{{3.90}} 
& \multicolumn{1}{c|}{{{1.49}}}  
& \multicolumn{1}{c|}{{{0.81}}}  
& \multicolumn{1}{c?}{{0.23}}               
& \multicolumn{1}{c|}{{{10.00}}}              
& \multicolumn{1}{c|}{{{0.18}}}               
& \multicolumn{1}{c|}{{\crosscheck}}        
& \multicolumn{1}{c|}{{\crosscheck}}        
& \multicolumn{1}{c?}{{\crosscheck}}        
& \multicolumn{1}{c|}{{\greencheck}}        
& \multicolumn{1}{c|}{{{0.50}}} 
& \multicolumn{1}{c|}{\cellcolor{myOrange}{{80.68}}} \\          

\cline{2-15} 
\multicolumn{1}{|c|}{{ }}                               
& \multicolumn{1}{c|}{}                                 
& \multicolumn{1}{c|}{{\textbf{DiCE}}}                           
& \multicolumn{1}{c|}{{0.78}}               
& \multicolumn{1}{c|}{{0.70}}               
& \multicolumn{1}{c|}{{8.42}}               
& \multicolumn{1}{c?}{{0.08}}               
& \multicolumn{1}{c|}{{1.25}}               
& \multicolumn{1}{c|}{{0.02}}               
& \multicolumn{1}{c|}{{\greencheck}}        
& \multicolumn{1}{c|}{{\greencheck}}        
& \multicolumn{1}{c?}{{\greencheck}}        
& \multicolumn{1}{c|}{{\crosscheck}}        
& \multicolumn{1}{c|}{\cellcolor{myBlue}{{0.80}}}               
& \multicolumn{1}{c|}{{0.08}}   \\          
\cline{3-15}

\multicolumn{1}{|c|}{\multirow{-6}{*}{{ \begin{tabular}[c]{@{}c@{}}\textbf{Covertype} \\  NumFeat: 10 \\ CatFeat: 44\\ EncFeat: 44 \end{tabular} }}} 
& \multicolumn{1}{c|}{\multirow{-2}{*}{\begin{tabular}[c]{@{}c@{}}\textbf{Neural} \\ \textbf{Network}\end{tabular}}} 
& \multicolumn{1}{c|}{{\textbf{Prototype}}}                      
& \multicolumn{1}{c|}{{3.99}}              
& \multicolumn{1}{c|}{{1.51}}               
& \multicolumn{1}{c|}{{0.81}}               
& \multicolumn{1}{c?}{\cellcolor{myOrange}{{0.25}}}               
& \multicolumn{1}{c|}{{10.00}}              
& \multicolumn{1}{c|}{\cellcolor{myOrange}{{0.18}}}               
& \multicolumn{1}{c|}{{\crosscheck}}        
& \multicolumn{1}{c|}{{\crosscheck}}        
& \multicolumn{1}{c?}{{\crosscheck}}        
& \multicolumn{1}{c|}{{\greencheck}}        
& \multicolumn{1}{c|}{{0.55}}               
& \multicolumn{1}{c|}{{27.23}}  \\          
\hline
\hline

\multicolumn{1}{|c|}{{ }}                   
& \multicolumn{1}{c|}{}     
& \multicolumn{1}{c|}{{\textbf{DiCE}}}                           
& \multicolumn{1}{c|}{\cellcolor{myBlue}{{0.89}}}               
& \multicolumn{1}{c|}{\cellcolor{myBlue}{{0.73}}}               
& \multicolumn{1}{c|}{\cellcolor{myBlue}{{1.05}}}               
& \multicolumn{1}{c?}{\cellcolor{myBlue}{{0.09}}}               
& \multicolumn{1}{c|}{\cellcolor{myBlue}{{1.57}}}               
& \multicolumn{1}{c|}{{\cellcolor{myBlue}{0.05}}}               
& \multicolumn{1}{c|}{{\greencheck}}        
& \multicolumn{1}{c|}{{\greencheck}}        
& \multicolumn{1}{c?}{{\greencheck}}        
& \multicolumn{1}{c|}{{\crosscheck}}        
& \multicolumn{1}{c|}{\cellcolor{myBlue}{{0.70}}}               
& \multicolumn{1}{c|}{\cellcolor{myBlue}{{0.04}}}   \\          
\cline{3-15}

\multicolumn{1}{|c|}{{ }}      
& \multicolumn{1}{c|}{\multirow{-2}{*}{\begin{tabular}[c]{@{}c@{}}\textbf{Decision} \\ \textbf{Tree}\end{tabular}}}  
& \multicolumn{1}{c|}{{\textbf{Prototype}}}                      
& \multicolumn{1}{c|}{{5.46}}              
& \multicolumn{1}{c|}{{1.63}}               
& \multicolumn{1}{c|}{{1.74}}               
& \multicolumn{1}{c?}{{0.44}}               
& \multicolumn{1}{c|}{{19.56}}              
& \multicolumn{1}{c|}{{0.59}}               
& \multicolumn{1}{c|}{{\crosscheck}}        
& \multicolumn{1}{c|}{{\crosscheck}}        
& \multicolumn{1}{c?}{{\crosscheck}}        
& \multicolumn{1}{c|}{{\greencheck}}        
& \multicolumn{1}{c|}{{0.80}}               
& \multicolumn{1}{c|}{{10.66}}  \\          
\cline{2-15}

\multicolumn{1}{|c|}{{ }}   
& \multicolumn{1}{c|}{}                                 
& \multicolumn{1}{c|}{{\textbf{DiCE}}}                           
& \multicolumn{1}{c|}{{1.72}}               
& \multicolumn{1}{c|}{{0.84}}               
& \multicolumn{1}{c|}{{1.08}}               
& \multicolumn{1}{c?}{{0.23}}               
& \multicolumn{1}{c|}{{3.58}}               
& \multicolumn{1}{c|}{{0.11}}               
& \multicolumn{1}{c|}{{\greencheck}}        
& \multicolumn{1}{c|}{{\greencheck}}        
& \multicolumn{1}{c?}{{\greencheck}}        
& \multicolumn{1}{c|}{{\crosscheck}}        
& \multicolumn{1}{c|}{\cellcolor{myBlue}{{0.67}}}               
& \multicolumn{1}{c|}{{0.05}}  \\           
\cline{3-15}

\multicolumn{1}{|c|}{{ }}   
& \multicolumn{1}{c|}{\multirow{-2}{*}{\begin{tabular}[c]{@{}c@{}}\textbf{Random} \\ \textbf{Forest}\end{tabular}}}  
& \multicolumn{1}{c|}{{\textbf{Prototype}}}                  
& \multicolumn{1}{c|}{\cellcolor{myOrange}{{5.60}}} 
& \multicolumn{1}{c|}{\cellcolor{myOrange}{{1.77}}}  
& \multicolumn{1}{c|}{\cellcolor{myOrange}{{1.62}}}  
& \multicolumn{1}{c?}{{0.41}}               
& \multicolumn{1}{c|}{\cellcolor{myOrange}{{15.00}}}              
& \multicolumn{1}{c|}{\cellcolor{myOrange}{{0.45}}}               
& \multicolumn{1}{c|}{{\crosscheck}}        
& \multicolumn{1}{c|}{{\crosscheck}}        
& \multicolumn{1}{c?}{{\crosscheck}}        
& \multicolumn{1}{c|}{{\greencheck}}        
& \multicolumn{1}{c|}{\cellcolor{myOrange}{{0.05}}} 
& \multicolumn{1}{c|}{\cellcolor{myOrange}{{98.97}}} \\          

\cline{2-15} 
\multicolumn{1}{|c|}{{ }}                               
& \multicolumn{1}{c|}{}                                 
& \multicolumn{1}{c|}{{\textbf{DiCE}}}                           
& \multicolumn{1}{c|}{{1.90}}               
& \multicolumn{1}{c|}{{1.02}}               
& \multicolumn{1}{c|}{{1.68}}               
& \multicolumn{1}{c?}{{0.17}}               
& \multicolumn{1}{c|}{{3.62}}               
& \multicolumn{1}{c|}{{0.11}}               
& \multicolumn{1}{c|}{{\greencheck}}        
& \multicolumn{1}{c|}{{\greencheck}}        
& \multicolumn{1}{c?}{{\greencheck}}        
& \multicolumn{1}{c|}{{\crosscheck}}        
& \multicolumn{1}{c|}{\cellcolor{myBlue}{{0.84}}}               
& \multicolumn{1}{c|}{{0.04}}   \\          
\cline{3-15}

\multicolumn{1}{|c|}{\multirow{-6}{*}{{\begin{tabular}[c]{@{}c@{}}\textbf{Road} \\ \textbf{Safety} \\  NumFeat: 29 \\ CatFeat: 3\\ EncFeat: 3  \end{tabular}}}} 
& \multicolumn{1}{c|}{\multirow{-2}{*}{\begin{tabular}[c]{@{}c@{}}\textbf{Neural} \\ \textbf{Network}\end{tabular}}} 
& \multicolumn{1}{c|}{{\textbf{Prototype}}}                      
& \multicolumn{1}{c|}{{5.32}}              
& \multicolumn{1}{c|}{{1.55}}               
& \multicolumn{1}{c|}{{2.08}}               
& \multicolumn{1}{c?}{\cellcolor{myOrange}{{0.40}}}               
& \multicolumn{1}{c|}{{20.00}}              
& \multicolumn{1}{c|}{\cellcolor{myOrange}{{0.61}}}               
& \multicolumn{1}{c|}{{\crosscheck}}        
& \multicolumn{1}{c|}{{\crosscheck}}        
& \multicolumn{1}{c?}{{\crosscheck}}        
& \multicolumn{1}{c|}{{\greencheck}}        
& \multicolumn{1}{c|}{{0.20}}               
& \multicolumn{1}{c|}{{27.14}}  \\          
\hline

\end{tabular}
}

\end{table}

\end{document}